\documentclass[journal]{IEEEtran}

\usepackage{booktabs} % For formal tables
\usepackage{svg}
\usepackage{amsmath}
\usepackage{amssymb}
\usepackage{algorithmic}
\usepackage{algorithm}
\usepackage{fancyhdr}

\usepackage[font=small,skip=1pt]{caption}

\usepackage{subfig}

\begin{document}
\title{Trading-off Accuracy and Energy of Deep Inference on Embedded Systems: A Co-Design Approach}
% \titlenote{Produces the permission block, and copyright information}
%\subtitle{Extended Abstract}
%\subtitlenote{The full version of the author's guide is available as
% \texttt{acmart.pdf} document}

\author{Nitthilan Kannappan Jayakodi, ~\IEEEmembership{Student Member, IEEE}, Anwesha Chatterjee, ~\IEEEmembership{Student Member, IEEE}, 

Wonje Choi,~\IEEEmembership{Student Member, IEEE}, Janardhan Rao Doppa,~\IEEEmembership{Member, IEEE}, 

Partha Pratim Pande,~\IEEEmembership{Senior Member, IEEE}
%<-this % stops a space
\thanks{This work was supported, in part by the US National Science Foundation (NSF) grants CNS-1564014 and CCF 1514269 and USA Army Research Office grant W911NF-17-1-04}
\thanks{Nitthilan K. J, Anwesha Chatterjee, Wonje Choi, J. R. Doppa, and P. P. Pande are with the School of Electrical Engineering and Computer Engineering, Washington State University, Pullman, WA 99163 USA (e-mail: [n.kannappanjayakodi, anwesha.chatterjee, wonje.choi, jana.doppa, pande]@wsu.edu).}% <-this % stops a space
\thanks{This article was presented in the International Conference on Hardware/Software Codesign and System Synthesis  (CODES + ISSS) and appears as part of the ESWEEK-TCAD special issue}

}

% \authornote{}
% \orcid{}
% \affiliation{
% %\institution{School of EECS, Washington State University, Pullman}
%   %\streetaddress{P.O. Box 1212}
%   %\city{Pullman}
%  % \state{WA, US}
%   %\postcode{43017-6221}
% }
% \email{{n.kannappanjayakodi, 
% anwesha.chatterjee, wonje.choi, jana.doppa, pande}@wsu.edu}

\IEEEoverridecommandlockouts
\IEEEpubid{\makebox[\columnwidth]{978-1-5386-5541-2/18/\$31.00~\copyright2018 IEEE \hfill} \hspace{\columnsep}\makebox[\columnwidth]{ }}
\maketitle
\IEEEpubidadjcol
% \vspace{-10.5ex}

\IEEEspecialpapernotice{(This article was presented in the International Conference on Hardware/Software Codesign and System Synthesis  (CODES + ISSS) and appears as part of the ESWEEK-TCAD special issue)}

\begin{abstract}
Deep neural networks have seen tremendous success for different modalities of data including images, videos, and speech. This success has led to their deployment in mobile and embedded systems for real-time applications. However, making repeated inferences using deep networks on embedded systems poses significant challenges due to constrained resources (e.g., energy and computing power). To address these challenges, we develop a principled co-design approach. Building on prior work, we develop a formalism referred as {\em Coarse-to-Fine Networks (C2F Nets)}  that allow us to employ classifiers of varying complexity to make predictions. We propose a principled optimization algorithm to automatically configure  C2F Nets for a specified trade-off between accuracy and energy consumption for inference. The key idea is to select a classifier on-the-fly whose complexity is proportional to the hardness of the input example: simple classifiers for easy inputs and complex classifiers for hard inputs. We perform comprehensive experimental evaluation using four different C2F Net architectures on multiple real-world image classification tasks. Our results show that optimized C2F Net can reduce the Energy Delay Product (EDP) by 27 to 60 percent with no loss in accuracy when compared to the baseline solution, where all predictions are made using the most complex classifier in C2F Net.
\end{abstract}

%\vspace{-0.3ex}


% <concept_id>10010147.10010257.10010293.10010294</concept_id>
% <concept_desc>Computing methodologies~Neural networks</concept_desc>
% <concept_id>10010583.10010662.10010674.10011722</concept_id>
% <concept_desc>Hardware~Chip-level power issues</concept_desc>
% <concept_id>10010520.10010553.10010560</concept_id>
% <concept_desc>Computer systems organization~System on a chip</concept_desc>
% <concept_significance>300</concept_significance>

% \ccsdesc[500]{Computing methodologies~Neural networks}
% \ccsdesc[500]{Hardware~Chip-level power issues}
% \ccsdesc[300]{Computer systems organization~System on a chip}

\begin{IEEEkeywords}
% \keywords{
Deep neural networks, Inference, Embedded systems, Approximate computing, Bayesian optimization, Hardware and software co-design
% }
\end{IEEEkeywords}

\IEEEpeerreviewmaketitle

% \vspace{-2.1ex}

\section{Introduction}

We are witnessing the rise of data-driven systems, where real-time predictions and decisions are being made based on models learned from large-scale data (e.g., text, images, speech, and sensor data). Deep learning --- a set of computational techniques to automatically extract patterns and useful features from raw data --- played a major role in this data revolution. Success stories of deep neural networks (DNNs) include achieving very high accuracy in image classification, speech recognition, and machine translation \cite{DL-Nature-2015}; deep reinforcement learning \cite{Deep-RL-Survey-2017}; and computers playing the game of GO against best human players (i.e., AlphaGo) \cite{AlphaGo-Nature-2016}. %The three main ingredients of these successes are: 1) advances in computing hardware to train large DNNs, 2) availability of large amount of labeled data, and 3) advances in DNN architectures and optimization strategies for robust training of large-scale models. 

In spite of the above successes, there are significant challenges for deploying trained DNNs to make predictions on edge devices (e.g., mobiles, Internet of Things, and Wearables) due to their constrained resources including energy and computing power. In practice, high accuracy is achieved by employing large DNNs, where making inference or predictions is computationally expensive and consumes a lot of energy. Unfortunately, this is not compatible with edge applications (e.g., robotics, smart health, and surveillance systems). As we discuss in related work section, most prior work has addressed these challenges using two different approaches from hardware and software perspective: 1) designing high-performance and energy-efficient hardware accelerators for performing inference, and 2) compressing large-scale DNNs with negligible loss in accuracy. In both these approaches the inference for {\em every} input example is made using a ``fixed'' computational process. They do not exploit the fact that hardness of inference varies from one example to another.

% \vspace{-1.8ex}

\begin{figure}[h]
\begin{minipage}{0.49\columnwidth}
    \includegraphics[width=\linewidth]{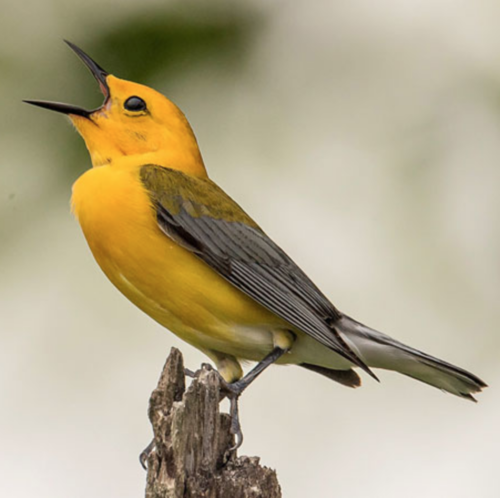}
    \end{minipage}
    \hspace{\fill} % note: no blank line here
    \begin{minipage}{0.49\columnwidth}
    \includegraphics[width=\linewidth]{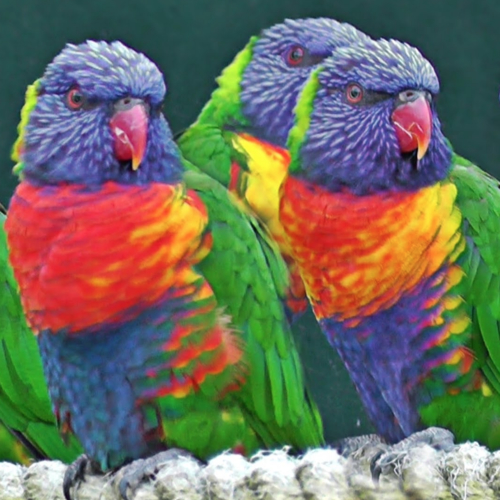}
    \end{minipage}
%     \vspace*{1cm} % vertical separation

%     \begin{minipage}{0.47\textwidth}
%     \includegraphics[width=\linewidth]{images/image_tile_2}
%     \end{minipage}
%     \hspace{\fill} % note: no blank line here
% \vspace{-1.2ex}
\caption{Example images to illustrate ``easy'' and ``hard'' inference problems. Left image is easy due to the existence of a single bird with clear background. Right image is hard due to the presence of multiple birds with complex textures.}  
\label{fig:easy-hard-images}
%Image Complexities (Bird) Left: Simple image (single bird, clear background) , Right: Complex image (multiple birds, complex textures} \label{fig:diff_image_types}
% \vspace{-2.0ex}
\end{figure}

In this paper, we develop a co-design approach to overcome the above drawback by performing ``adaptive computing''. We build on the general principle of coarse-to-fine computation \cite{C2FFaceDet},TagDyn and formalize a model that we refer as {\em Coarse-to-Fine Networks (C2F Nets)}, which is a generalization of the recently proposed conditional deep learning (CDL) model \cite{CondDeepLearn}. C2F Nets allow us to employ classifiers of varying complexity depending on the hardness of input examples. The key idea is to {\em learn} to select simpler (coarser) networks to classify ``easy'' examples and complex (finer) networks to classify ``hard'' examples. Figure~\ref{fig:easy-hard-images} illustrates easy and hard inputs for image classification task. We also provide design principles to guide the design of good C2F nets for a given real-world classification task. We propose a novel and principled stage-wise optimization approach to automatically configure C2F Nets for a specified trade-off between accuracy and energy consumption of inference on a target hardware platform. Our overall co-design approach using C2F Nets is complementary to prior approaches based on hardware and software optimization including the recent work on CDL \cite{CondDeepLearn}. 

%The key principle behind C2F Nets approach can potentially enable sustainable computing using a network of diverse computing systems ranging from edge devices to data centers as needed. Consider the following two extreme cases: 1) Computation for every input happens on edge device, and 2) Sending every input to a data center and getting back the output. We can employ the coarse-to-fine computation principle to process easy inputs on edge devices and harder inputs on computing systems with appropriate resources to achieve high-performance and energy-efficiency. 

We performed comprehensive experiments using four different C2F Nets for image classification tasks. We optimized the C2F Nets for different trade-offs between accuracy and energy consumption of inference on a target hardware platform. Our results show that many input examples can be classified correctly using simple networks and complex networks are used only for hard examples. Our optimized C2F Net reduces the Energy Delay Product (EDP) by 27 to 60 percent with no loss in accuracy when compared to the baseline solution, where all predictions are made using the most complex network in C2F Nets.

% \vspace{1.0ex}

\noindent {\bf Contributions.} We make three main contributions in this paper.

\begin{enumerate}
\item Formalize the coarse-to-fine network (C2F Net) model by generalizing the recent work on CDL. This model allows to perform adaptive computing to make inference on input examples of varying hardness. We also provide a methodology to guide the design of C2F networks for practitioners based on fine-grained energy analysis.
\item Develop a principled optimization algorithm to automatically configure C2F Nets for a specified trade-off between accuracy and energy on a target hardware platform.
\item Comprehensive experimental evaluation of four C2F Net architectures on real-world image classification to demonstrate the effectiveness of overall approach.
\end{enumerate}

\section{Related Work}

%The main constraints for efficient deployment of deep learning applications on embedded systems are memory requirements and computation and energy cost. Memory access and data movements are the main bottlenecks for DNN processing in embedded systems and accounts for significant energy cost \cite{Eyeriss}.  Several works have been proposed that address these constraints both at the architectural level as well as software level \cite{Survey_v.Sze}.

\begin{figure*}[h]
\begin{minipage}{\textwidth}
\begin{minipage}{0.70\textwidth}
    \includegraphics[trim={10 35 10 10},clip,width=\linewidth]{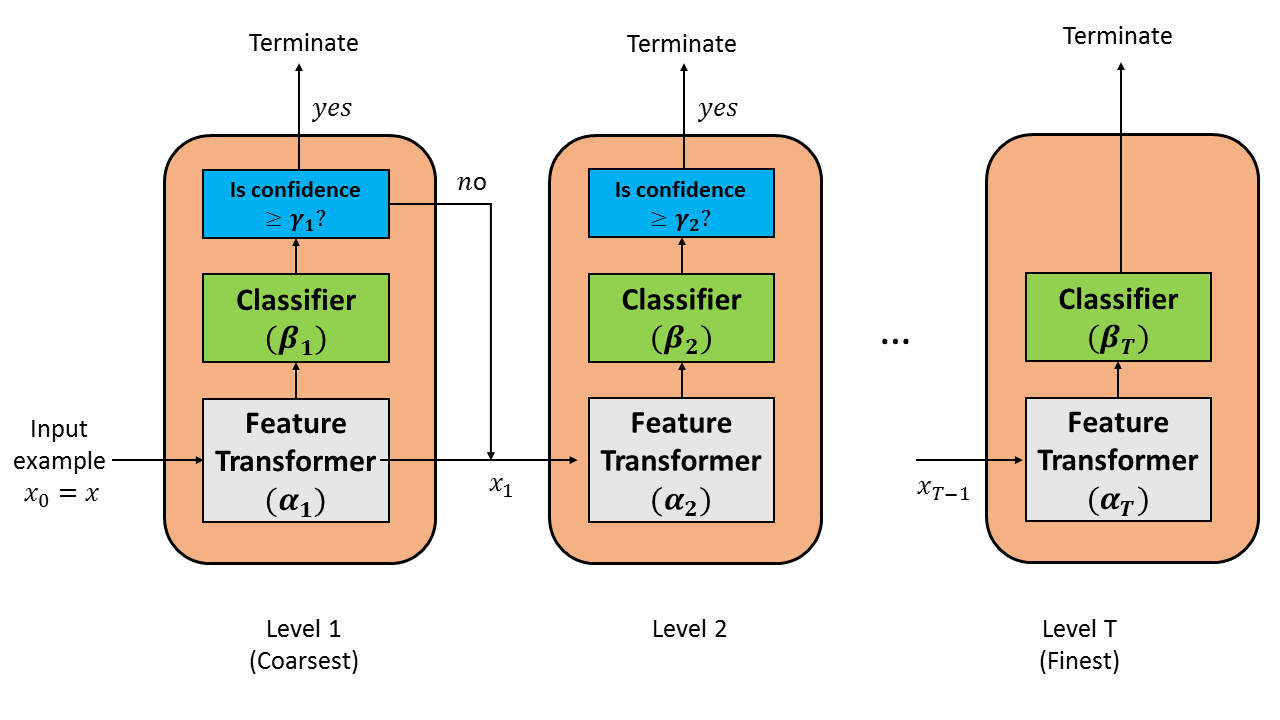}
    \centering
    %\small (a) An abstract C2F net model with $T$ levels from coarsest to finest network, where each level consists of a feature transformer, a classifier, and a confidence threshold to decide when to terminate.
    \end{minipage}
    \hspace{\fill} % note: no blank line here
    \begin{minipage}{0.29\textwidth}
    \includegraphics[width=\linewidth]{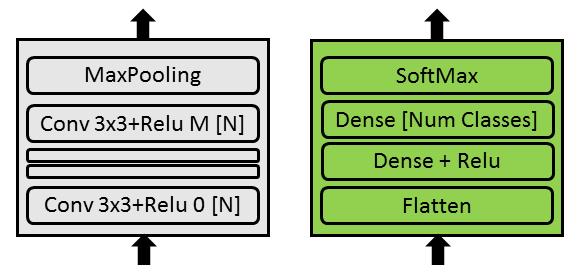}
    \centering
    \small Building blocks for C2F nets. The feature transformer block $[MxN]$ has a series of convolution layers followed by a max-pooling layer. $M$ and $N$ represents the number of convolution layers and number of filters in each layer. The classifier block has a series of dense or fully connected layers followed by soft-max layer.
    \end{minipage}
    \hspace{\fill} % note: no blank line here
    %\begin{minipage}{0.60\textwidth}
    %\includegraphics[trim={10 35 10 10},clip,width=\linewidth]{images_block_diagram/c2f_net_a_blk_diag}
    %\centering
    %\small (c) Example C2F net A with three levels. At Level 1, we have 2 convolution layers with 64 filters each. At level 2, we have 2 convolution Layers (in addition to those in level 1) with 128 filters each. At level 3, we have 2 additional convolution Layers with 192 filters each.
    %\end{minipage}
% \vspace{-1.0ex}    
\caption{Illustration of Coarse-to-Fine (C2F) networks model.An abstract C2F net model with $T$ levels from coarsest to finest network, where each level consists of a feature transformer, a classifier, and a confidence threshold to decide when to terminate.} 
\label{fig:C2F-Net-abstract-figure}
\vspace{-2.5ex}
\end{minipage}
\end{figure*}

Main challenges for deploying deep learning applications on embedded systems include high resource requirements in terms of memory, computation, and energy. 
% Memory access and data movements are the main bottlenecks for making inference using DNNs \cite{Eyeriss}.
%on embedded systems and accounts for significant energy cost 
Prior work has addressed these challenges using methods based on hardware, software, and hardware/software co-design \cite{Survey_v.Sze}. %We discuss various approaches for these three categories below. 

\vspace{0.5ex} 

\noindent {\bf Hardware Approaches.} Architecture level methods include design of specialized hardware targeting DNN computations using the data-flow knowledge \cite{Eyeriss}, placement of memory closer to computation units \cite{TETRIS}, and on-chip integration of memory and computation \cite{DaDianNao}. Apart from the constraint of having an application-specific hardware system, most of these hardware technologies also have an overhead of analog to digital conversion \cite{Survey_v.Sze}.

\vspace{0.5ex}

\noindent {\bf Software Approaches.} To address the challenges in general purpose CPUs and GPUs, following software-level approaches are studied. Reducing the precision of weights and activations is one such approach \cite{DBLP:journals/corr/ZhouYGXC17}. 
%It involves quantization of the data and mapping into small number of levels thereby reducing the number of bits. 
Fixed point representation of weights and activations dynamically varying across different layers is also employed to reduce the energy and area cost of the network \cite{FPweights}. In \cite{DBLP:journals/corr/LaiSC17}, authors show that using floating-point for weights and fixed-point for activations leads to reduction of power consumption by 50\% and reduction in model size by 36\%.  Another software level approach is pruning and compression of large-scale DNNs \cite{KnowledgeDist}. The sparsity in features and activations after non-linear operations like ReLU is exploited to compress the models in \cite{Eyeriss}. To address significant redundancy in weights during training the network, the authors in \cite{DBLP:journals/corr/HanPTD15} employ pruning technique on the weights based on their magnitudes. 
%Weights with small magnitude are pruned, and rest of the weights are fine-tuned to maintain the accuracy. 
While magnitude-based pruning of weights reduces the computation cost, it does not directly address the energy cost. To make the model more energy efficient, an ``energy-aware'' pruning mechanism is proposed in \cite{Energy-Aware}, where pruning of weights is performed layer-by-layer based on the knowledge of energy hungry layers. This method is shown to be 1.74 times more energy-efficient when compared to the weight-magnitude based pruning.  %Another compression approach is to learn a simpler network from a trained complex network via knowledge transfer .

A finer-level software approach is to improve the network architecture. The inception module \cite{Inception} performs a 2D-convolution considering two 1D convolutions considering the 2-D filters to be completely separable. The Xception \cite{Xception} and Mobilenets \cite{MobileNets} architectures also employ depth-wise separable convolutions similar to inception model, but perform an additional point-wise convolution, i.e., a 1x1 convolution at the end to combine the outputs of the depth-wise convolutions. This method %of factorizing the convolution layer into separable filters followed by a pointwise combining process 
helps in reducing the computation and model size.

%Another way of defining a network model is to derive a simpler network from a trained complex network by mechanism of knowledge transfer \cite{KnowledgeDist}. In this process a compressed network is gained with an accuracy that it couldn’t have achieved if trained directly with the same dataset. 

A recent paper \cite{ICNN-DATE2018} proposed an iterative CNN (ICNN) approach to make multi-stage predictions for images. The key idea is to apply a two-stage wavelet transform to produce multiple small-scale images and train separate models to process them progressively to make multi-stage predictions. Compared to ICNN approach, our C2F nets approach has following advantages: 1) C2F nets is more principled and general solution. We can easily instantiate the framework for different DNN architectures; and 2) Our optimization approach is very efficient and allows to automatically tune C2F nets for different energy-accuracy trade-offs for a target hardware. 

\vspace{0.5ex}

\noindent {\bf Hardware and Software Co-Design Approaches.} Many of the above approaches involve design of hardware guided by software level optimization techniques. \cite{FPweights} proposes the design of a compiler specific to FPGA that analyzes the structure and parameters of convolutional neural networks, and generates modules to improve the throughput of the system. In \cite{EIE}, authors propose the design of Energy Inference Engine (EIE) that deploys pruned DNN models. In \cite{Cnvlutin}, the knowledge of compressed sparse-weights guides the hardware to read weights and perform multiply-and-accumulate (MAC) computations only for the nonzero activations, thereby reducing the energy cost by 45\%.   

\vspace{0.5ex}

\noindent {\bf Coarse-to-Fine and Conditional Inference Approaches.} Coarse-to-Fine or cascaded computation is a general principle that has a long history in computer science and machine learning \cite{C2FFaceDet,TagDyn}. The application of this general principle to a given problem is non-trivial, and the exact details and algorithmic procedures vary from one application to another including our co-design approach for inference with deep neural networks.

Cascaded CNN architecture for face detection \cite{ConvFaceDetect} progressively prune areas of the image that are not likely to contain a face. The key differences with our work are as follows: 1) In our C2F architecture, the features computed in the previous level are reused in the next level to construct more complex features, whereas the CNNs used in \cite{ConvFaceDetect} do not reuse any features; and 2) The approach proposed in \cite{ConvFaceDetect} do not allow to optimize the cascade for a specified trade-off between speed and accuracy of face detection.

Prior work on coarse-to-fine deep kernel networks \cite{DeepKernel} tries to combine multiple input kernels in a recursive manner to achieve a complex kernel and associated representation of the data. This paper makes very strict assumptions and incorporates these assumptions directly into the overall learning approach: 1) It can only work for binary classification tasks; 2) It assumes that the number of negative examples are significantly higher than positive examples. The learning approach in \cite{DeepKernel} for training coarser networks is tuned to binary classification and assigns very high-costs to misclassifying positive examples to incorporate this assumption; and 3) There are no thresholds to configure the architecture to trade-off speed and accuracy of prediction. Similarly, there is no optimization methodology to optimize the architecture for a specified trade-off.

The high-level architecture of branch CNN (BCNN) \cite{FastBranch} is similar in style to our C2F model, but there are some significant differences: 1) BCNN uses linear classifier after each layer, whereas we do not; and 2) The approach in \cite{FastBranch} doesn't allow to configure the BCNN for any specified trade-off between speed and accuracy. This method is also very {\em ad hoc} as there is no well-defined optimization formulation and optimization methodology.

Recent paper on conditional deep learning (CDL) model \cite{CondDeepLearn} is the closest related work to ours. However, there are some significant differences between our work and this paper: 1) CDL model employs a single threshold value at all the levels, whereas we employ different thresholds for different levels. When the number of levels are more, different thresholds for different levels will achieve better pareto-optimal solutions. As discussed later, our experimental results on CIFAR10 and CIFAR100 clearly show this difference; 2) In their training approach, they only train using a subset of the input examples that are not filtered by the previous level. This will likely cause over-fitting and may not work well on harder tasks. We train all the intermediate classifiers on all the training examples to improve their generalization. Additionally, the approach in \cite{CondDeepLearn},  need to train all the different levels for each confidence threshold, which is very time-consuming. In our case, we train classifiers at different levels only once and tune the thresholds for different levels jointly conditioned on the trained features and classifiers; and 3) We look at the co-design problem of configuring a software application to run on a specific hardware architecture for a specified trade-off between accuracy and energy. Our optimization approach based on Bayesian Optimization principles to configure the C2F net to trade-off accuracy and energy is significantly different and allows us to achieve better pareto-optimal solutions. Our experimental evaluation is much more comprehensive (CIFAR10, CIFAR100, and MNIST datasets with different C2F Nets). We perform in-depth energy analysis of different layers to provide principles to design C2F nets and also fine-grained analysis to understand why the approach works better.

\section{Coarse-to-Fine Networks (C2F Nets)}

In this section, we describe the details of our coarse-to-fine networks (C2F Nets) model. %that allows us to adaptively select a classifier of appropriate complexity depending on the hardness of input example. 
In what follows, we provide the motivation, formal model of C2F Nets along with some examples, and how to make predictions given a fully specified C2F Net model.

%- We are inspired from Viola and Jones Face detector and Coarse-to-Fine principle of making predictions.

%- A pipeline of ``T'' blocks, where each block `i' consists of the following components.

% $\alpha$ -i = parameters of feature network

% $\beta$ -i = parameters of classifier

% $\gamma$ -i = confidence threshold

% - An abstract picture with $\alpha$, $\beta$, $\gamma$ variables, and exit decisions. Different choices of confidence variables.

% - Describe how predictions are made using this C2F architecture.

% - A concrete example coarse-to-fine network for image classification.

%- Illustrative results to show the accuracy and energy consumption of classifiers $\beta$-1 (coarsest) to $\beta$-T (finest) for the above example.

%- Show three real-world images (either from our image dataset or else where) that show the varying hardness: easy, hard, and very hard from a naked eye perspective.

%- Our goal is to optimize C2F Net so that it uses appropriate classifer for a specified trade-off for accuracy and energy 

% \vspace{-2.5ex}
\subsection{Motivation}

Simple (shallow) DNNs are energy-efficient, but their accuracy is low. On the other hand, many state-of-the-art DNNs that achieve very high accuracy are highly complex (deep) and consume significant amount of energy. We are motivated by the fact that many input examples are {\em easy} and can be classified correctly using simple DNNs and only a small fraction of {\em hard} inputs would need complex DNNs. We illustrate this observation using two examples. First, Figure~\ref{fig:easy-hard-images} shows bird images corresponding to easy and hard cases. Second, Figure~\ref{fig:base_perf} shows the accuracy of different DNNs with varying complexity. We can see that the accuracy improvement is small when we move from simple to complex DNNs, whereas their energy consumption grows significantly. This corroborates our hypothesis. Therefore, our goal is automatically select simple networks for easy inputs and complex networks for hard inputs. This mechanism will allow us to achieve high accuracy with significantly less energy consumption when compared to the baseline approach, where we make predictions for all input examples using the most complex DNN.

% \begin{figure*}[h]
% \includegraphics[width=\textwidth]{images_block_diagram/C2F-Net-abstract-figure}
% \caption{Coarse to Fine Architecture} \label{fig:C2F-Net-abstract-figure}
% \end{figure*}

% \vspace{-3.3ex}
\subsection{C2F Nets Model}

A C2F Net model $\mathcal{NN}$ with $T$ levels can be seen as a sequence of networks stacked  in the order of increasing complexity as shown in Figure~\ref{fig:C2F-Net-abstract-figure}. Each level $i$ of the model contains three key elements.

% \vspace{0.8ex}

 {\bf 1) Feature transformation function} in the form of a network of layers that takes the features computed in previous level $x_{i-1}$ as input and produces more complex features $x_i$. Parameters of the feature function are denoted by $\alpha_i$. Figure~\ref{fig:C2F-Net-abstract-figure} and Figure~\ref{fig:c2f_type_blk_diag} provide illustrative examples for feature transformation functions.

% \vspace{0.8ex}

 {\bf 2) Classifier} takes the features computed by feature transformation function $x_i$ and produces a probability distribution over all candidate labels. Parameters of the classifier are denoted by $\beta_i$. Figure~\ref{fig:C2F-Net-abstract-figure} provides an illustration of the classifier block.

% \vspace{0.8ex}

{\bf 3) Confidence threshold.}
%\cite{ConfThresh} 
Given a predicted probability distribution over classification labels, we can estimate the confidence of the classifier in a variety of ways \cite{ConfThresh} including: a) {\em Maximum probability}, where we take the probability of the highest score label; b) {\em Entropy} over the probability of all candidate labels; c) {\em Margin} denoted by the difference in probability scores between best and second-best label. For a given confidence type, the confidence threshold $\gamma_i$ is employed to determine if the classifier is confident enough in its prediction to terminate and return the predicted label.

% \vspace{-3.0ex}
\subsection{Inference in C2F Nets}

Given a C2F Net with all its parameters ($\alpha$, $\beta$, and $\gamma$) fully specified, inference computation for a new input example $x$ is performed as follows (see Algorithm~\ref{alg:c2f-inference}). We sequentially go through the C2F Net starting from level 1. At each level $i$, we make predictions using the feature transformation functions parameterized by $\alpha_i$ and intermediate classifiers $\beta_i$ to compute the probability distribution of classification labels $\hat{P_i}(y)$. We estimate the confidence parameter $t_i$ from $\hat{P_i}(y)$ and predicted output $\hat{y}$ as the label with highest probability score. If estimated confidence in prediction $t_i$ meets the threshold $\gamma_i$ or we reach the final level $T$, we terminate and return the predicted output $\hat{y}$ for the given input $x$.

% \vspace{-2.0ex}
\section{Optimization Approach for C2F Nets}

In this section, we describe a principled optimization algorithm to automatically configure C2F Nets for a specified trade-off between accuracy and energy on a target hardware platform. 

\begin{algorithm}[H]
\caption{Inference in Coarse-to-Fine Networks}
\textbf{Input}: $x = $ input example; 
$\mathcal{NN} = $ C2F network with $T$ levels;
$\alpha_1, \alpha_2, \cdots, \alpha_T = $ parameters of feature transformers; 
$\beta_1, \beta_2, \cdots, \beta_T = $ parameters of intermediate classifiers; 
$\gamma_1, \gamma_2, \cdots, \gamma_{T-1} = $ confidence threshold values \\
\begin{algorithmic}[1]
\STATE $x_0 \leftarrow x$; and level $i \leftarrow$ 1
\STATE TERMINATE $\leftarrow$ \texttt{False}
\WHILE{TERMINATE == \texttt{False}}
\STATE $x_i \leftarrow$ \textsc{Transform-Features}($x_{i-1}$, $\alpha_i$)
\STATE $\hat{P_i}(y) \leftarrow$ \textsc{Intermediate-Classifier}($x_i$, $\beta_i$)
\STATE Prediction $\hat{y} \leftarrow$ label with highest probability from $\hat{P_i}(y)$
\STATE Compute confidence $t_i$ using $\hat{P_i}(y)$
\IF{$i$ == $T$ OR $t_i \geq \gamma_i$}
\STATE TERMINATE $\leftarrow$ \texttt{True}
\ELSE
\STATE $i \leftarrow i+1$ // Continue with next level
\ENDIF
\ENDWHILE
\STATE \textbf{return} predicted class label $\hat{y}$
\end{algorithmic}
\label{alg:c2f-inference}
\end{algorithm}

% \vspace{-5.5ex}

{\bf Problem Setup.} Without loss of generality, let us assume that we are given a C2F Net $\mathcal{NN}$ with $T$ levels. The parameters of $\mathcal{NN}$ can be divided into three parts: 1) Parameters of feature transformation functions at different levels $\alpha$ = ($\alpha_1, \alpha_2,\cdots,\alpha_T$); 2) Parameters of intermediate classifiers at different levels $\beta$ = ($\beta_1, \beta_2,\cdots,\beta_T$); and 3) Confidence thresholds at different levels $\gamma$ = ($\gamma_1, \gamma_2,\cdots,\gamma_{T-1}$) that decide the complexity of classifier employed for a given input example. We are provided with a training set $D_{train}$ and validation set $D_{val}$ of classification examples drawn from an unknown target distribution $\mathcal{D}$, where each classification example is of the form $(x, y^*)$, where $x$ is the input (e.g., image) and $y$ is the output class (e.g., image label). We are also given a target hardware platform $\mathcal{H}$ for which the C2F model $\mathcal{NN}$ need to be optimized for. We can measure the energy consumption and accuracy of inference a candidate $\mathcal{NN}$ model using the hardware $\mathcal{H}$. Suppose $\mathcal{O}(\mathcal{NN}, \mathcal{H}, \lambda)$= $\mathbb{E}_{(x,y^*) \sim \mathcal{D}}$ $\lambda \cdot$ \textsc{Error}($\mathcal{NN}$, $\mathcal{H}$) +  (1- $\lambda)\cdot $\textsc{Energy}($\mathcal{NN}$, $\mathcal{H}$) stands for the expected error and energy trade-off achieved over the target distribution of classification examples $\mathcal{D}$ when the coarse-to-fine network model $\mathcal{NN}$ is executed with parameters $\alpha$, $\beta$, $\gamma$ on the target hardware platform $\mathcal{H}$.
For a specified trade-off parameter $\lambda \in \left[0, 1\right]$, our goal is to find the best parameters $\alpha$, $\beta$, $\gamma$ of coarse-to-fine network $\mathcal{NN}$ to achieve the specified trade-off between error and energy for the target platform $\mathcal{H}$.

% \vspace{-3.0ex}
%
\begin{equation}
{(\alpha^*, \beta^*, \gamma^*) = \mbox{arg}\min_{\alpha, \beta, \gamma} \mathcal{O}(\mathcal{NN}, \mathcal{H}, \lambda)}
\label{opt-problem}
\end{equation}

Since we do not have access to the distribution $\mathcal{D}$, we employ the training and validation set to find the best parameter values. For example, to evaluate a candidate parameter configuration over the validation set, we do the following. We compute the normalized error and normalized energy with respect to the setting where all predictions are made using the largest network over all the input examples in the validation set. We plug the normalized error and energy in the objective $\mathcal{O}$ to evaluate the candidate parameter configuration.

% \vspace{-2.0ex}
\subsection{Stagewise Optimization Algorithm}

% Temporary argument. Need to be refined.
The optimization problem posed in Equation~\ref{opt-problem} is extremely challenging to solve due to complex interactions between the parameter values  ($\alpha$, $\beta$, $\gamma$) on the optimization objective $\mathcal{O}(\mathcal{NN}, \mathcal{H}, \lambda)$. Generally, C2F Model $\mathcal{NN}$ would be trained using back-propagation via stochastic gradient descent (SGD) optimization \cite{BackProp}. Specifically, the Objective requires the energy consumption of coarse-to-fine model $\mathcal{NN}$ on the target hardware platform $\mathcal{H}$, where it is executed. Applying the standard back-propagation training algorithm would require us to estimate the energy for every iteration of the training step, which would make the training procedure impractical. 
% Specifically, the accuracy and energy consumption of  coarse-to-fine model $\mathcal{NN}$ depends on the target hardware platform $\mathcal{H}$, where it is executed. So we cannot use standard back-propagation training algorithm to find the parameters that optimize the objective $\mathcal{O}$. 

To overcome these challenges, we leverage the structure in this optimization problem to develop an efficient stagewise optimization algorithm (see Algorithm~\ref{alg:stagewise}), where the parameters $\alpha$, $\beta$, and $\gamma$ are optimized sequentially one after another. By splitting it into multiple stages, we decouple the dependency of the training procedure for $\alpha$ and $\beta$ parameters from the energy measurement step. First, we learn the parameters $\alpha$ to optimize the feature transformers at different levels. Second, conditioned on the found $\alpha$, we learn the parameters $\beta$ to optimize the intermediate classifiers at different levels. In both the stages, we optimize only for the prediction accuracy independent of the energy. Third, conditioned on the found $\alpha$ and $\beta$, we find the best values of confidence thresholds $\gamma$ to optimize the objective $\mathcal{O}$. We describe the details of these three optimization procedures in the following subsections. 

% \vspace{-2.5ex}

\begin{algorithm}[H]
\caption{Stagewise Optimization of C2F Nets}
\textbf{Input}: $D_{train} = $ training set of classification examples; 
$D_{val} = $ validation set of classification examples;
$\mathcal{NN} = $ C2F Network architecture with $T$ levels;
$\mathcal{H} = $ target hardware platform;
$\lambda = $ parameter to trade-off accuracy and energy; \\
%\textbf{Output}: $\hat{\gamma}_1, \hat{\gamma}_2, \cdots, \hat{\gamma}_{T-1}$ = best thresholds from $\mathcal{G}$ 
\begin{algorithmic}[1]
\STATE $\alpha$ = $(\alpha_1,\alpha_2,\cdots,\alpha_T) \leftarrow$ \textsc{Train-Features}($D_{train}$)
\STATE $\beta$ = $(\beta_1,\beta_2,\cdots,\beta_T) \leftarrow$ \textsc{Train-Classifiers}($D_{train}$, $\alpha$)
\STATE $\gamma$ = $(\gamma_1,\gamma_2,\cdots,\gamma_{T-1}) \leftarrow$ \textsc{Optimize-Thresholds}($D_{val}$, $\mathcal{H}$, $\lambda$, $\alpha$, $\beta$, $\mathcal{G}$), where $\mathcal{G}$ is the joint space of $T$-1 confidence thresholds
\STATE \textbf{return} the optimized parameters of C2F Net $\alpha$, $\beta$, and $\gamma$
\end{algorithmic}
\label{alg:stagewise}
\end{algorithm}

% \vspace{-4.5ex}

\subsection{Optimizing Feature Transformers}

To obtain the parameters $\alpha$ corresponding to feature transformation functions at different levels, we train the finest (most complex) C2F Net as follows. We minimize the cross-entropy loss over training data $D_{train}$ using a SGD training procedure. Cross-entropy loss (aka log loss) measures the performance of a classification model whose output is a probability value between 0 and 1. Cross-entropy loss increases as the predicted probability diverges from the actual label. SGD is a stochastic approximation of the gradient descent optimization and is a iterative method for minimizing cross-entropy loss via back-propagation.

% \vspace{-2.5ex}

\begin{algorithm}[H]
\caption{Training Feature Transformers}
\textbf{Input}: $D_{train} = $ training set of classification examples \\
\begin{algorithmic}[1]
\STATE $\alpha_1, \alpha_2, \cdots, \alpha_T \leftarrow $ Train the finest (most complex) classifier in C2F Net to minimize the cross-entropy loss via back propagation algorithm
\STATE \textbf{return} parameters of feature transformers $\alpha_1, \alpha_2, \cdots, \alpha_T$
\end{algorithmic}
\label{alg:alpha}
\end{algorithm}

% \vspace{-4.5ex}

\subsection{Optimizing Intermediate Classifiers}

Our approach to optimize the parameters of intermediate classifiers (i.e., $\beta$) is inspired by the transfer learning approach employed in deep learning. The key idea is to leverage the pre-trained model for a source task to quickly learn a model for a relevant target task. In our case, the source task corresponds to learning parameters $\alpha$ for the finest classifier as described above and the target task corresponds to learning parameters $\beta_i$ for intermediate classifier at level $i \in \left\{1,2,\cdots,T\right\}$. Intuitively, we want to minimize the energy consumption as much as possible by making correct classification decisions with high confidence using simple (coarser) classifiers. Therefore, we fix the parameters $\alpha_1,\alpha_2,\cdots,\alpha_i$ for feature transformation functions and learn the parameters $\beta_i$ by minimizing the cross-entropy loss over the training data $D_{train}$. 

% \vspace{-2.5ex}

\begin{algorithm}[H]
\caption{Training Intermediate Classifiers}
\textbf{Input}: $D_{train} = $ training set of classification examples; 
$\alpha_1, \alpha_2, \cdots, \alpha_T = $ parameters of feature transformers \\ 
\begin{algorithmic}[1]
\FOR{each level $i$ = $1, 2,\cdots, T$ of C2F Net}
\STATE $\beta_i \leftarrow $ Train a classifier with feature transformers $\alpha_1, \alpha_2,\cdots,\alpha_i$ to minimize the cross-entropy loss
\ENDFOR
\STATE \textbf{return} parameters of intermediate classifiers $\beta_1, \beta_2, \cdots, \beta_{T}$
\end{algorithmic}
\label{alg:beta}
\end{algorithm}

% \vspace{-3.5ex}
% \vspace{-1.1ex}

\subsection{Optimizing Confidence Thresholds}

In this section, we describe our approach for finding the best thresholds $\gamma$=($\gamma_1, \gamma_2, \cdots, \gamma_{T-1}$) for the specified trade-off between accuracy and energy consumption of inference with C2F networks.

% \vspace{0.75ex}

{\bf Problem Formulation.} Suppose the domain of each confidence threshold $\gamma_i$ at each level $i \in \left\{1,2,\cdots,T-1\right\}$ is $\left[lb, ub\right]$. For example, if we employ probability of highest scoring label as our confidence parameter, then $lb$=0 and $ub$=1. Let $\mathcal{G}$ denote the joint space of candidate confidence threshold assignments, where $g \in \mathcal{G}$ is a candidate assignment to all the $T$-1 threshold parameters $\gamma_1, \gamma_2,\cdots,\gamma_{T-1}$. We can evaluate the objective $\mathcal{O}(\mathcal{NN}, \mathcal{H}, \lambda)$= $\mathbb{E}_{(x,y^*) \sim \mathcal{D}}$ $\lambda \cdot$ \textsc{Error}($\mathcal{NN}$, $\mathcal{H}$) +  (1- $\lambda)\cdot $\textsc{Energy}($\mathcal{NN}$, $\mathcal{H}$) for any $g \in \mathcal{G}$ by running the corresponding C2F net model $\mathcal{NN}$ on the validation set of classification examples $D_{val}$ (i.e., expectation is approximated with a finite sample). Our goal is to find the best confidence thresholds $g^* \in \mathcal{G}$ that will lead to the highest objective value $\mathcal{O}(g)$.

\begin{figure}[H]
\includegraphics[width=\columnwidth]{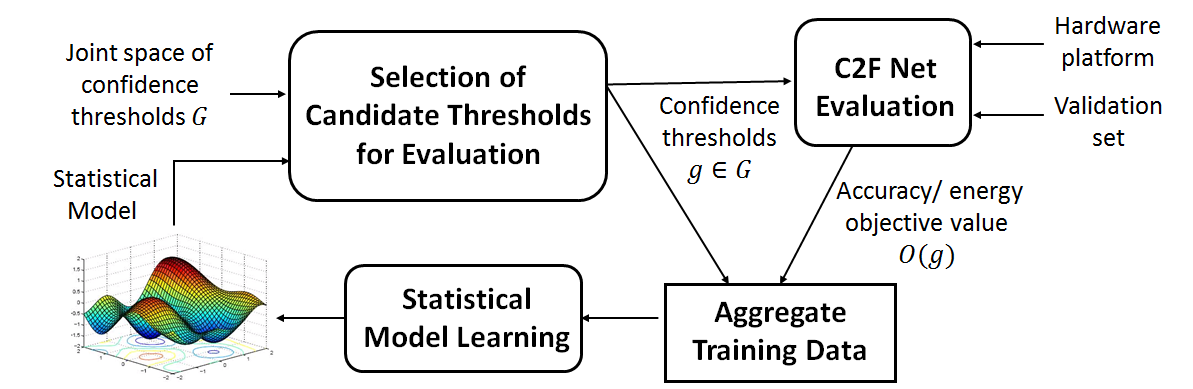}
\caption{High-level overview of Bayesian Optimization approach for selecting the best confidence threshold values to configure a given C2F Net on a target hardware platform.} \label{fig:BO-for-C2F-Net}
\end{figure}
% \vspace{-3.0ex}

{\bf Bayesian Optimization Approach.} We propose to solve the above problem using the Bayesian Optimization (BO) framework \cite{BO} that is known to be very efficient in solving global optimization problems using black-box evaluations of the objective function (Figure \ref{fig:BO-for-C2F-Net}). BO algorithms can be seen as sequential decision-making processes that select the next candidate input to be evaluated to quickly direct the search towards optimal inputs by trading-off exploration and exploitation at each search step. By iteratively selecting a candidate input for evaluation and learning a statistical model based on the observed input and output pairs, BO approach can quickly move towards high-quality inputs; this significantly reduces the number of objective function evaluations during the optimization process. 

In what follows, we describe the three key elements of the general BO framework as applicable for our problem:
% \vspace{0.05ex}

 {\bf 1) A Statistical Model} of the true function $\mathcal{O}(g)$ by placing a {\em prior} over the space of functions. {\em Gaussian Process (GP)} \cite{GP} is the most commonly used prior due to its generality and uncertainty quantification ability. A GP over a space $\mathcal{G}$ is a random process from $\mathcal{G}$ to $\mathbb{R}$. It is characterized by a mean function $\mu : \mathcal{G} \rightarrow \mathbb{R}$ and a covariance or kernel function $\kappa : \mathcal{G} \times  \mathcal{G} \rightarrow \mathbb{R}$. {\em Radial kernels} are typically employed as prior covariance functions of GPs. Using a radial kernel means that the prior covariance can be written as $\kappa(g, g') = \kappa_{0}\phi(\left\|g - g'\right\|)$ and depends only on the distance between $g$ and $g'$. The scale parameter $\kappa_{0}$ captures the magnitude the function could deviate from $\mu$. $\phi$ is a decreasing function with $\phi(0)=1$. If a function $F$ is sampled from $\mathcal{GP}(\mu,\kappa)$, then $\mathcal{O}(g)$ is distributed normally $\mathcal{N}(\mu(g), \kappa(g,g))$ for all $g \in \mathcal{G}$. %Suppose that we are given $t$ experimental evaluations $\mathcal{D}_t = \left\{(g_i, \mathcal{O}(g_i))\right\}_{i=1}^{t}$ from this GP, where $g_i \in \mathcal{G}$ is a candidate confidence threshold assignment and $\mathcal{O}(g_i)$ is the corresponding objective value of accuracy/energy trade-off. Then the posterior process with respect to $\mathcal{D}_t$ is also a GP with mean $\mu_t$ and covariance $\kappa_t$.

% \vspace{0.25ex}

{\bf 2) An Acquisition Function} \textsc{Af} to score the utility of evaluating a candidate input based on the current statistical model. \textsc{Af} need to trade-off exploration and exploitation based on the predictions of the statistical model. Exploitation corresponds to selecting candidate inputs for which the statistical model is highly confident (high mean) and exploration corresponds to selecting candidate inputs for which the statistical model is not confident (high variance). We employ the popular Upper Confidence Bound (UCB) rule as acquisition function. UCB function is defined as $UCB_{t+1}(x) = \mu_{t}(x) + \sqrt{\kappa_{t+1}} \sigma_{t}(x)$, where $\mu_t$ and $\sigma_t$ are the posterior mean and standard deviation of the GP conditioned on $\mathcal{D}_t$ (Algorithm \ref{alg:gamma}). $\mu_t$ encourages exploitation, $\sigma_t$ encourages exploration, and $\kappa_{t+1}$ controls the trade-off. 

% \vspace{0.25ex}

{\bf 3) An Optimization Procedure} to select the the best scoring candidate input according to \textsc{Af}. We employ the popular DIvided RECTangles (DIRECT) approach to select the input $g \in \mathcal{G}$ to be queried in each iteration guided by the statistical model.

In each iteration, BO algorithm calls the optimizer to select the next input $g \in \mathcal{G}$ to evaluate the objective $\mathcal{O}(g)$, and update the statistical model using the aggregate training data from past evaluations (see Algorithm \ref{alg:gamma}).

\begin{algorithm}[H]
\caption{Finding Best Thresholds via Bayesian Optimization}
\textbf{Input}: $D_{val} = $ validation set of classification examples; 
$\mathcal{H} = $ target hardware platform;
$\lambda = $ parameter to trade-off accuracy and energy;
$\alpha_1, \alpha_2, \cdots, \alpha_T = $ parameters of feature transformers; 
$\beta_1, \beta_2, \cdots, \beta_T = $ parameters of intermediate classifiers; $\mathcal{G} = $ joint space of confidence thresholds \\
%\textbf{Output}: $\hat{\gamma}_1, \hat{\gamma}_2, \cdots, \hat{\gamma}_{T-1}$ = best thresholds from $\mathcal{G}$ 
\begin{algorithmic}[1]
\STATE $\mathcal{D}_0 \leftarrow$ small number of input-output pairs; and $t \leftarrow$ 0
\REPEAT
\STATE Learn the GP model: $\mathcal{M}_t \leftarrow$ \textsc{Learn-GP}($\mathcal{D}_t$)
\STATE Compute the next input to query:  \\ $g_{t+1} \leftarrow \mbox{arg}\,max_{g \in \mathcal{G}} \, \textsc{Af}(\mathcal{M}_t, g)$
\STATE Query objective function $\mathcal{O}$ at $g_{t+1}$ to get $\mathcal{O}(g_{t+1})$
\STATE Aggregate the data: $\mathcal{D}_{t+1} \leftarrow \mathcal{D}_{t} \cup \left\{(g_{t+1}, \mathcal{O}(g_{t+1}))\right\}$
\STATE $t \leftarrow t+1$
\UNTIL{convergence or maximum iterations}
\STATE ${g}_{best} \leftarrow \mbox{arg}\,max_{g_t} \, \mathcal{O}(g_{t})$
\STATE \textbf{return} best found thresholds ${g}_{best}$=$\gamma_1, \gamma_2, \cdots, \gamma_{T-1}$
\end{algorithmic}
\label{alg:gamma}
\end{algorithm}

% \vspace{-2.5ex}

% \vspace{-1.5ex}

% \vspace{-2.0ex}

\section{C2F Nets Design Methodology}

In this section, we describe our methodology to design C2F networks from a given base network architecture (i.e., complex DNN). Although the methodology is generally applicable to different neural network architectures, we use convolutional neural networks for image classification as a running example for illustrative purposes. 
%Our goal is to come up with a sequence of networks that cover different accuracy and energy trade-offs for a specific hardware platform. 
%In what follows, we first describe our hardware setup, followed by our design methodology guided by energy analysis of different layers in the base neural network architecture.

% \vspace{-2.5ex}

\subsection{Hardware Setup}

All our experiments were performed by deploying DNN models on an \texttt{ODROID-XU4} board \cite{odroidboard}. \texttt{ODROID-XU4} is an Octa-core heterogeneous multi-processing system with ARM BigLittle architecture, which is very popular in current mobile devices. The BigLittle architecture consists of asymmetrical multi-core system, where cores are clustered into two groups based on the power and performance modes: 1) the power hungry (Big) mode; and 2) the battery saving slower processor mode (Little).  The \texttt{ODROID-XU4} board employs ARM Cortex-A15 Quad 2GHz CPU as the Big Cluster and Cortex-A7 Quad 1.4GHz CPU as the Little Cluster. The board boots up Ubuntu 16.04LTS with ODROID's Linux kernel version 3.10.y. We execute DNN models with Caffe-HRT \cite{Caffe-HRT} that employs Caffe framework \cite{Caffe} with ARM Compute Library to speed up deep learning computations. We employ SmartPower2 \cite{SP2} to measure power. We compute the average power over the total execution time.

% \vspace{-2.5ex}

\subsection{Fine-Grained Energy Analysis of Base DNN}

\noindent {\bf Base deep neural network (DNN) architectures.} We consider deep convolutional networks for solving image classification task. These architectures consist of different layers including Convolution, Fully Connected (FC) or Dense, Max Pooling, ReLU, Soft-Max, and Batch-Norm. For ease of understanding, we divide the network into {\em feature transformation block} and {\em classifier block}. The feature transformer block is the top part of the network and consists of convolution and max-pooling layers. It generates the input features for classifier block. The classifier block is the bottom part of the network  and consists of fully connected layers followed by a soft-max layers. It predicts the final classification label for the input image. Refer Figure \ref{fig:C2F-Net-abstract-figure} for further details.
We considered four different base DNNs with differing complexity to solve image classification task for our fine-grained energy analysis: {\em Base Network A and C} contains six convolution layers and {\em Base Network B and D} consists of ten convolution layers (a slight variant of VGG network \cite{VGG}). However, we present our analysis and results for network A noting that the results are similar for all the other networks.

%\vspace{-1.0ex}

\begin{figure}[h]
\centering
\includegraphics[trim={10 0 45 40},clip,width=0.75\columnwidth]{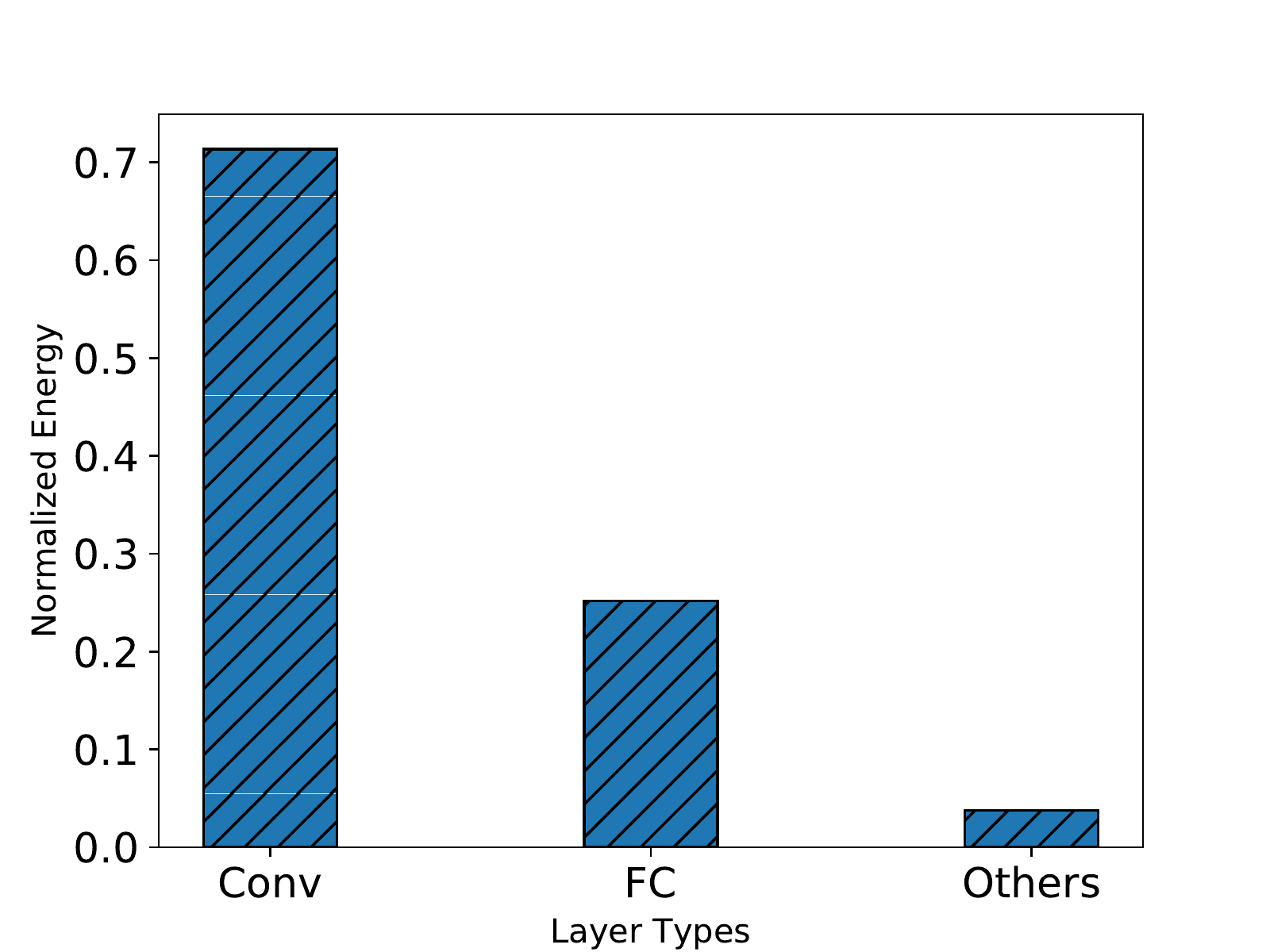}
% \vspace{-1.0ex}
\caption{Energy distribution of different layers for base network A.}%: Convolution layers, fully connected layers, and all other layers.}
\label{fig:diff_layer_energy_net_a}
% \vspace{-2.5ex}
\end{figure}

To understand the influence of a DNN architecture on the energy consumption of a real-world application, we analyzed the energy consumption per layer for all base networks to classify a single image. Our high-level energy analysis shows that convolution layers consume the most energy followed by fully connected layers. Figure~\ref{fig:diff_layer_energy_net_a} shows the distribution of energy consumed by convolution layers, fully connected layers, and all other layers for network A. In what follows, we perform fine-grained analysis for convolution and fully connected layers separately. 

\begin{figure}[h]
\centering
	%left bottom right top 
   \includegraphics[trim={15 20 45 40},clip,width=0.75\columnwidth]{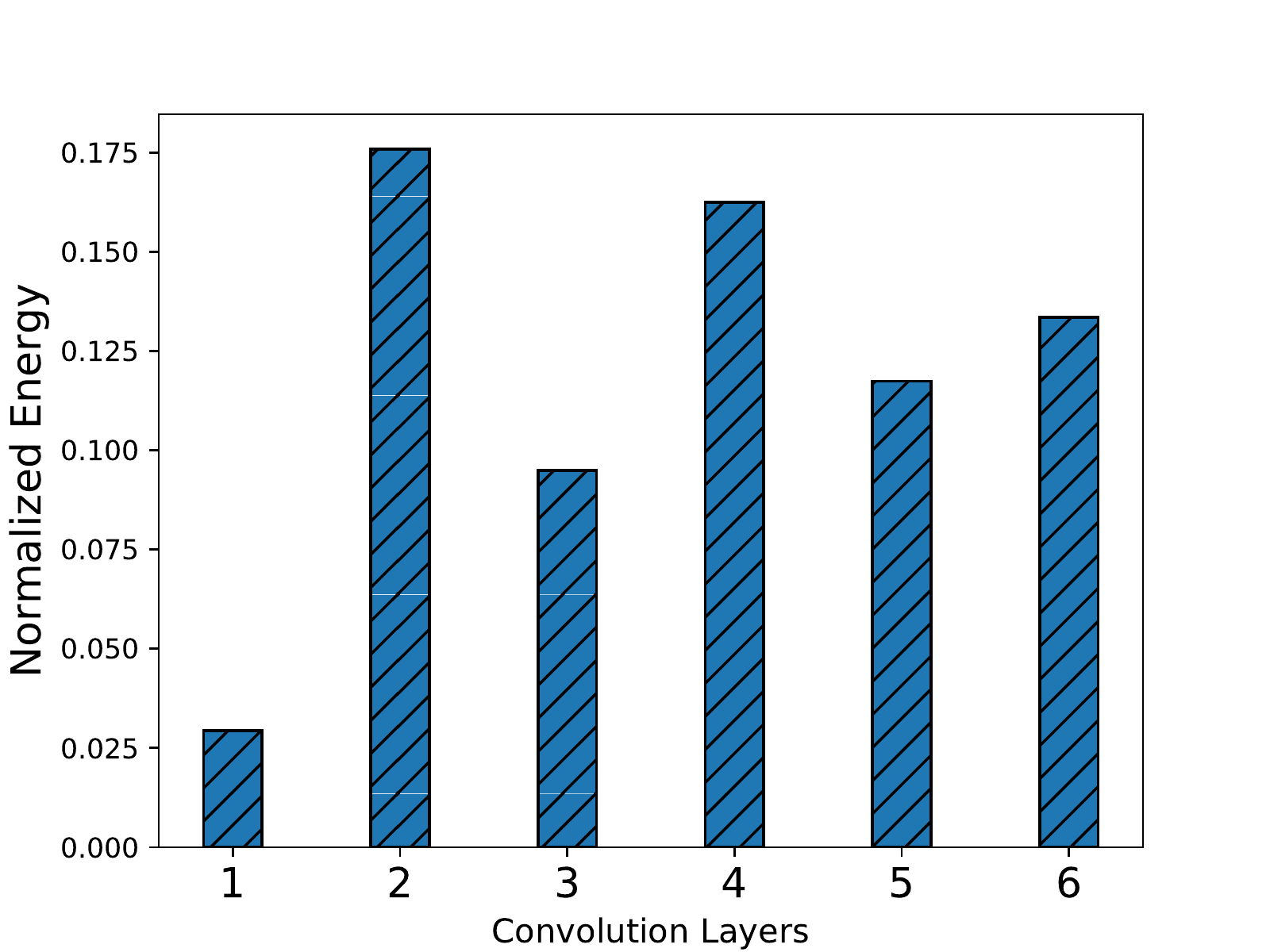}
	\caption{Energy consumption at different convolution layers for base network A.  Layer 1 is the closest to the input layer. All energy values are normalized with respect to total energy consumed for inference of one image.}
  \label{fig:conv_energy_net_a}
\end{figure}

% \begin{minipage}{0.010\linewidth}
% \end{minipage}
% \hspace{.009\linewidth}
\begin{figure}[h]
\centering
%left bottom right top 
\includegraphics[trim={15 20 45 40},clip,width=0.75\columnwidth]{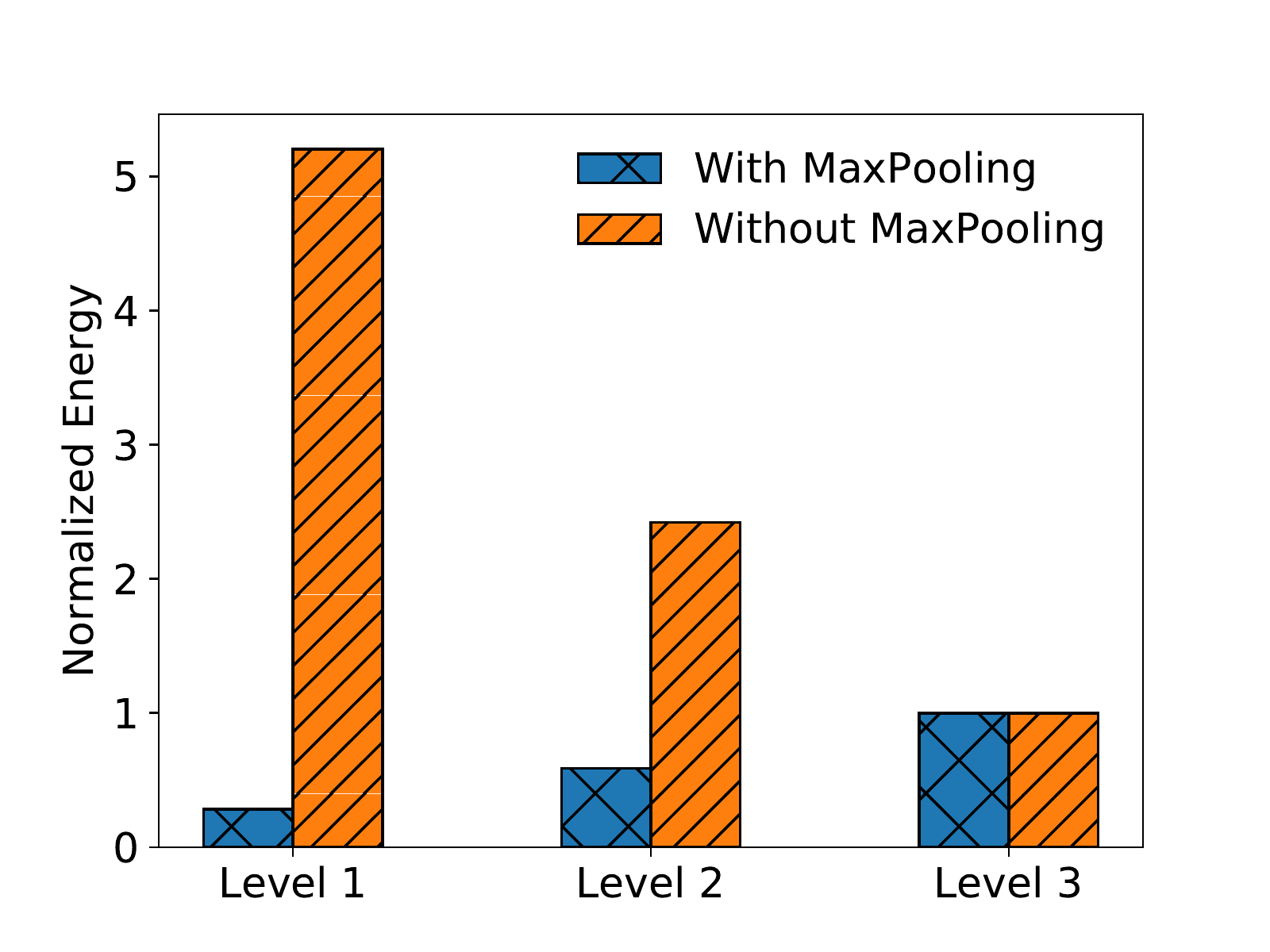}
    \caption{Energy consumption of FC layer for level 1, level 2, level 3 of C2F Net A. The effect of adding an extra max-pooling layer to reduce the energy. Energy values normalized w.r.t FC layer energy of the last level 3.}
    \label{fig:effect_of_fc_layer_a}
\end{figure}
% \vspace{0.5ex}

\noindent {\bf Energy analysis of convolution layers.} The energy consumed by a convolution layer depends on the following elements: a) the size of the filter, b) image dimensions over which the filter is applied, and c) the number of filters. For all the base networks (A, B, C, and D), the number of convolution filters increases with the depth of the layer. For example, in base network A, the number of filters for increasing depths are 64, 128, and 192 respectively. 

%i.e. in case of Base Net A it is 64, 128 and 192 and Base Net B it is 64, 128, 256, 512 respectively [\ref{fig:c2f_type_blk_diag}].

To design C2F Nets, we want to construct relatively simpler networks from base network for different trade-offs in accuracy and energy consumption. For these simpler (coarser) networks, we want to reduce the energy consumption by varying the values of the above-mentioned three key elements. We fix the size of the filter to 3x3 across all the layers for all the base networks. Therefore, our design of coarser networks is guided by reducing the number of convolution layers, number of filters per layer, and input image dimension. Figure~\ref{fig:conv_energy_net_a} shows the distribution of energy consumption (normalized with respect to the total energy consumed by the base network) across different convolution layers of the base network A. Therefore, by progressively combining different number of convolution layers, we can generate different coarser networks of progressively increasing energy consumption.

\noindent {\bf Energy analysis of fully connected layers.} Fully connected or dense layers consume the most energy after convolution layers. Dense layer is part of the classifier block of our base DNNs. The classifier block generally has two or more dense layers. Out of them, the first dense layer is usually the largest because it is obtained by flattening the features generated by the immediate convolution layer (just one Max Pooling layer) and is proportional to square of the image width. %For a image with dimension $d$, the number of features from convolution layer is proportional to $d^2$.
Hence, the first FC layer consumes significantly more energy than the rest.

%The Dense Layer is part of the Classifier block of our networks. The Classifier block usually has two or more dense layers. Of these, the first dense layer is usually the largest because it is obtained by flattening the features generated (Feature Transformer block) by the convolution layer which is proportional to square of the image width. For a image dimension of N, the number of features of the convolution layer is proportional to N\^2. Hence, among the different FC layers the first FC layers consumes the significant energy [as shown in the figure - not sure we are adding this figure].

% \begin{figure}[h]
%    \includegraphics[trim={0 0 45 40},clip,width=0.75\columnwidth]{images_design_principle/conv_energy_net_a}
% \caption{Distribution of energy consumption by different convolution layers for base network A. Convolution layer 1 is the closest to the input layer. All energy values are normalized with respect to total energy consumed for inference of one image.}
% \label{fig:conv_energy_net_a_net_b}
% \vspace{-4.5ex}
% \end{figure}

% \vspace{-2.0ex}

\subsection{Designing C2F Networks}

Our goal is to construct networks of varying complexity that show different trade-offs in accuracy and energy to be part of C2F Net. 

\noindent {\bf Convolution layer.} From the above discussion, we can construct networks that consume different amounts of energy by combining different number of convolution layers (acts as a feature transformer block for each level $i$) followed by a classifier for that level. For example, we construct a C2F Net for base network A with three levels as following. Level 1 (coarsest network) consists of two convolution layers as the feature block followed by a classifier block. Level 2 consists of four convolution layers (two extra convolution layers in addition to those from level 1). Level 3 corresponds to the full base network (finest network). Therefore, with increasing levels, we have progressively more convolution layers resulting in better accuracy at the expense of energy.

\noindent {\bf Fully connected layer.} While designing C2F networks, we add classifiers at different depths of convolution layer for each level. At small depth, the image width is large (due to fewer max-pooling layers) for convolution. Hence, large number of features are fed to the classifier block. This increases the energy consumption of fully connected layers in the classifier blocks for coarser levels. Figure~\ref{fig:effect_of_fc_layer_a} shows the energy consumptions across the different levels of C2F Net corresponding to base network A as described above without an extra max-pooling layer. 

We observe that the energy consumed by FC layer of level 1 and level 2 is significantly larger than the energy consumed by the FC layer of Finest Layer (level 3), which is not desirable.  Therefore, to match the classifier block energy across different levels, we add an extra max-pooling layer for each of the classifier block for coarser networks (level 1 and level 2) such that the FC input dimension and the corresponding energy is similar across all the levels. Figure\ref{fig:effect_of_fc_layer_a} shows that the energy consumption with the extra max-pooling layer is comparable across all levels. 

In summary, a practitioner can employ the above design principles to design coarse-to-fine networks that provide different trade-offs between accuracy and energy consumption.
% \vspace{-1.5ex}

\section{Experiments and Results}

In this section, we first describe the experimental setup and then discuss results along different dimensions of optimized C2F Nets.

% \vspace{-1.5ex}

\subsection{Experimental Setup}

% \vspace{0.8ex}

{\bf Hardware Setup.} We employ the hardware platform described in Section 5 to perform inference using different networks to measure the accuracy and energy consumption of inference process.

% \vspace{0.8ex}

{\bf Image classification task and Training data.} We consider a image classification task with 10 different classes. We employ the CIFAR10, CIFAR100 and MNIST \cite{cifar10} \cite{mnist} image dataset with a 4:1:1 split for training (40000), validation (10000), and testing (10000) to train and test our C2F Nets approach. The input image dimension is 32x32x3 for the CIFAR10 and CIFAR100 datasets while it is 32x32x1 for the MNIST data set. The number of classes to be predicted is 10 for CIFAR10 and MNIST while it is 100 for CIFAR100. We employed the CIFAR10 dataset to train our adaptive C2F nets A and B; MNIST dataset for C2F net C; and CIFAR100 dataset for C2F net D

% \vspace{0.8ex}

 {\bf Energy and accuracy trade-off objective $\mathcal{O}$.} Recall that the training of C2F Nets is based on the objective 
$\mathcal{O}(\mathcal{NN}, \mathcal{H}, \lambda)$= $\mathbb{E}_{(x,y^*) \sim \mathcal{D}}$ $\lambda \cdot$ \textsc{Error}($\mathcal{NN}$, $\mathcal{H}$) +  (1- $\lambda)\cdot $\textsc{Energy}($\mathcal{NN}$, $\mathcal{H}$) where $\mathcal{H}$ is the hardware platform and $\mathcal{NN}$ is the C2F Net. Since accuracy and error are complementary, we have presented all our results in terms of accuracy and energy for the ease of exposition. Accuracy is measured as the fraction of input images whose labels are predicted correctly. Energy consumption of C2F Nets is measured in terms of normalized energy delay product (EDP). EDP = $\sum$P $\Delta$t $\cdot$ T, where $\Delta$t is the time interval at which we record the power P and T is the total execution time. As power is measured at a regular interval we simply calculate EDP as EDP = $P_{avg}$ $\cdot$ $T^{2}$ . This value is normalized with EDP of the base network. We vary the value of $\lambda$ from 0 to 1 to get C2F Nets optimized for different trade-offs between accuracy and energy. 
% EDP is defined as the product of energy consumed and execution time for performing inference using C2F Nets on a given image.
% \vspace{0.8ex} 

% Nitthilan: C2F Net A and Net B figure here.
\begin{figure}[h]
\centering
\begin{minipage}{0.5\columnwidth}
\includegraphics[width=\linewidth]{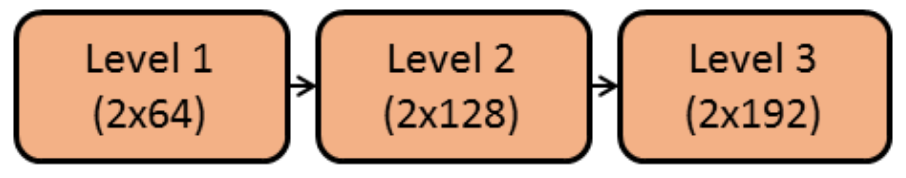}
    \centering
    \small (a) C2F Net A/Net C
\end{minipage}
\begin{minipage}{\columnwidth}
\begin{minipage}{0.495\columnwidth}
    \includegraphics[width=\linewidth]{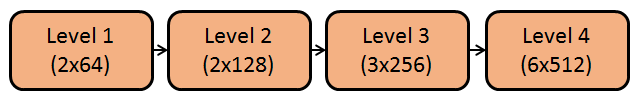}
    \centering
    \small (b) C2F Net B
\end{minipage}
\begin{minipage}{0.495\columnwidth}
    \includegraphics[width=\linewidth]{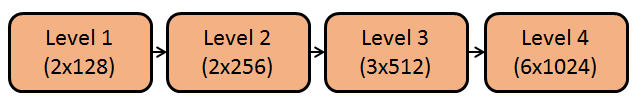}
    \centering
    \small (c) C2F Net D
\end{minipage}
\end{minipage}

\caption{C2F Nets definition in terms of the notation in figure \ref{fig:C2F-Net-abstract-figure}.} 
%(Top) C2F Net A (Bottom) C2F Net B. As described in figure \ref{fig:C2F-Net-abstract-figure}, C2F Net A has 64, 128 and 192 filters at Levels 1,2,3 with 2 convolution layers each. C2F Net B has 64, 128, 256 and 512 filters at Levels 1,2,3,4 with 2 layers in Level 1,2 and 3 layers in Level 3,4. } 
\label{fig:c2f_type_blk_diag}
% \vspace{-2.5ex}
\end{figure}

 {\bf C2F Networks.} We employ four C2F networks in our experimental evaluation. Figure~\ref{fig:c2f_type_blk_diag} shows the high-level architecture of C2F Net A and C with three levels and C2F Net B and D with four levels using the notations introduced in Figure~\ref{fig:c2f_type_blk_diag}. With Net A and Net B we demonstrate the effect of different architectures on the same CIFAR10 dataset. With Net C and Net D we analyze the impact on the performance for a simpler and complex dataset like MNIST and CIFAR100 respectively.

% \vspace{0.8ex}

 {\bf Confidence threshold types.} We experimented with all three types of confidence computation including {\em Max probability}, {\em Margin}, and {\em Entropy}. We observed that performance is almost same for all the three types. Therefore, we present all our results using Max probability as the confidence type.
%Unless otherwise mentioned all the presented results employ the maximum probability as the confidence type.

% \vspace{0.8ex} 

 {\bf Offline training of C2F Nets.} Recall that we need to find the parameters $\alpha$, $\beta$, and $\gamma$ for the specified trade-off objective using training and validation set of classification examples. For obtaining the parameters $\alpha$ and $\beta$, we employ the backpropagation training algorithm using RMS prop optimizer with learning rate of 0.0001 and a decay set to $1e^{-6}$. We employ a batch size of 128 and run training for 200 epochs with sufficient data augmentation including horizontal flips, width and height shifts. For obtaining the confidence thresholds, we employ the Bayesian Optimization (BO) approach with Gaussian kernel and UCB rule as the acquisition function. We use five evaluations of the objective $\mathcal{O}$ randomly for initialization. We run BO until convergence or a maximum of 100 iterations. We train C2F Nets for different values of $\lambda$ to get different trade-offs between energy and accuracy.

% \section{Results}
% In this section we discuss in detail the various implications of the approach we took, specifically the effect on prediction accuracy, energy consumption, prediction time across the two different Network A and B.

% \vspace{-2.0ex}

\subsection{Results and Discussion}

In this section, we present the results of our optimized C2F Nets and compare them along different dimensions.

\begin{figure}[h]
\begin{minipage}{\linewidth}
\begin{minipage}{0.495\linewidth}
	%left bottom right top 
    \includegraphics[trim={0 20 45 40},clip, width=\linewidth]{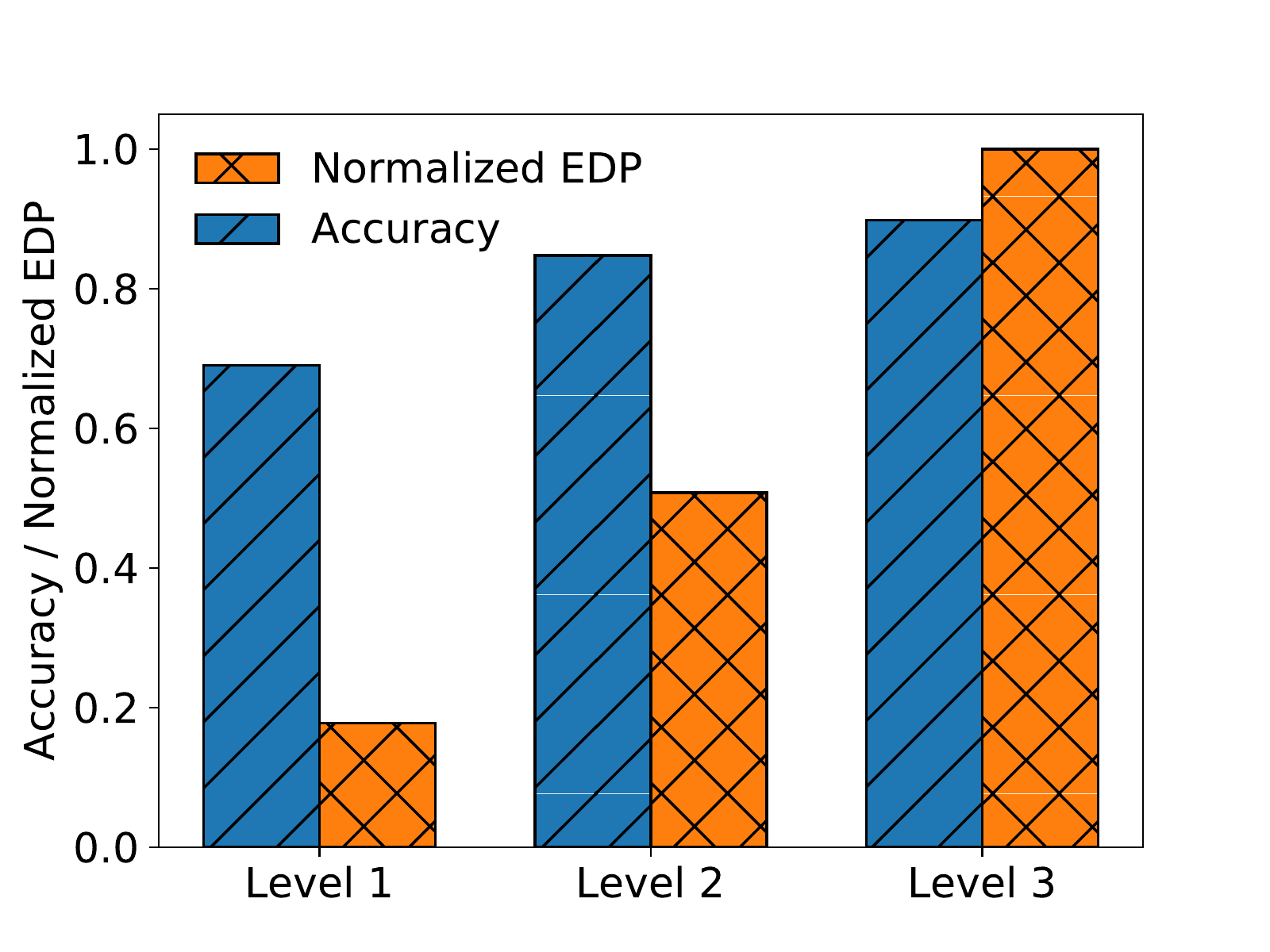}
    \centering
    \small (a) C2F Net A
    \end{minipage}
%     \hspace{\fill} % note: no blank line here
    \begin{minipage}{0.495\linewidth}
    \includegraphics[trim={0 20 45 40},clip, width=\linewidth]{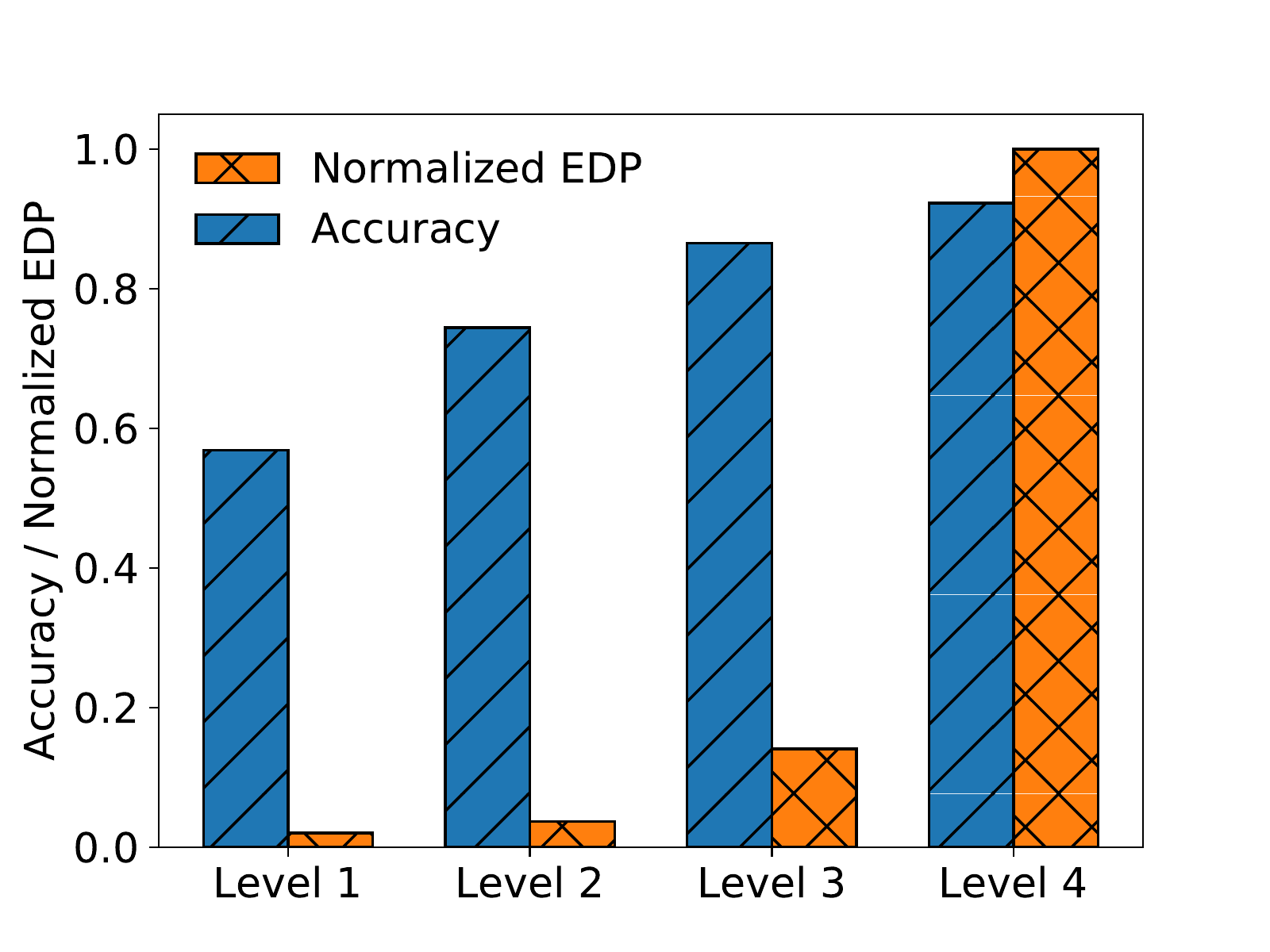}
%     \caption{c2f b}
	\centering
    \small (b) C2F Net B
    \end{minipage}
    \begin{minipage}{0.495\linewidth}
    \includegraphics[trim={0 20 45 40},clip, width=\linewidth]{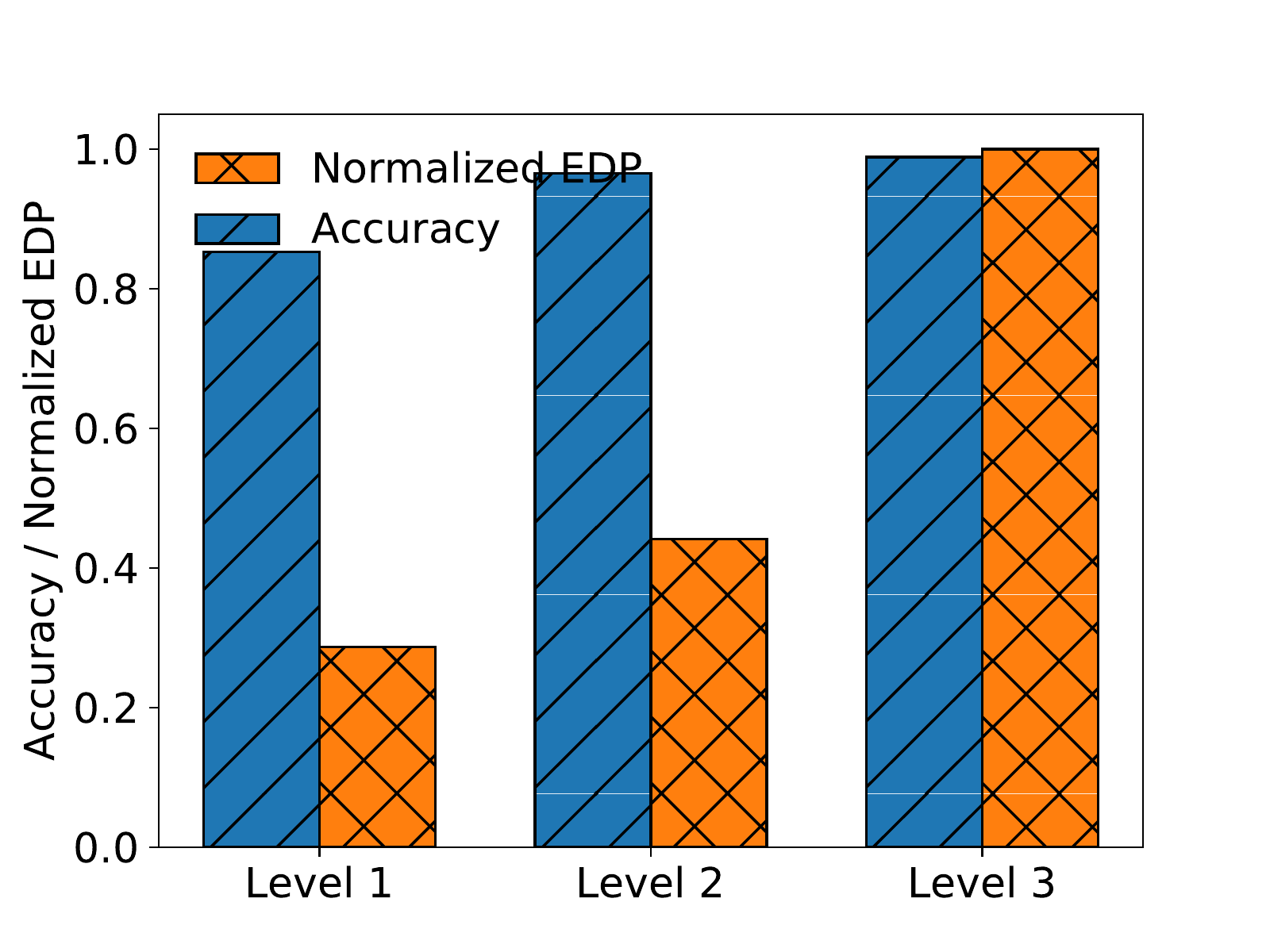}
%     \caption{c2f b}
	\centering
    \small (c) C2F Net C
    \end{minipage}
    \begin{minipage}{0.495\linewidth}
    \includegraphics[trim={0 20 45 40},clip, width=\linewidth]{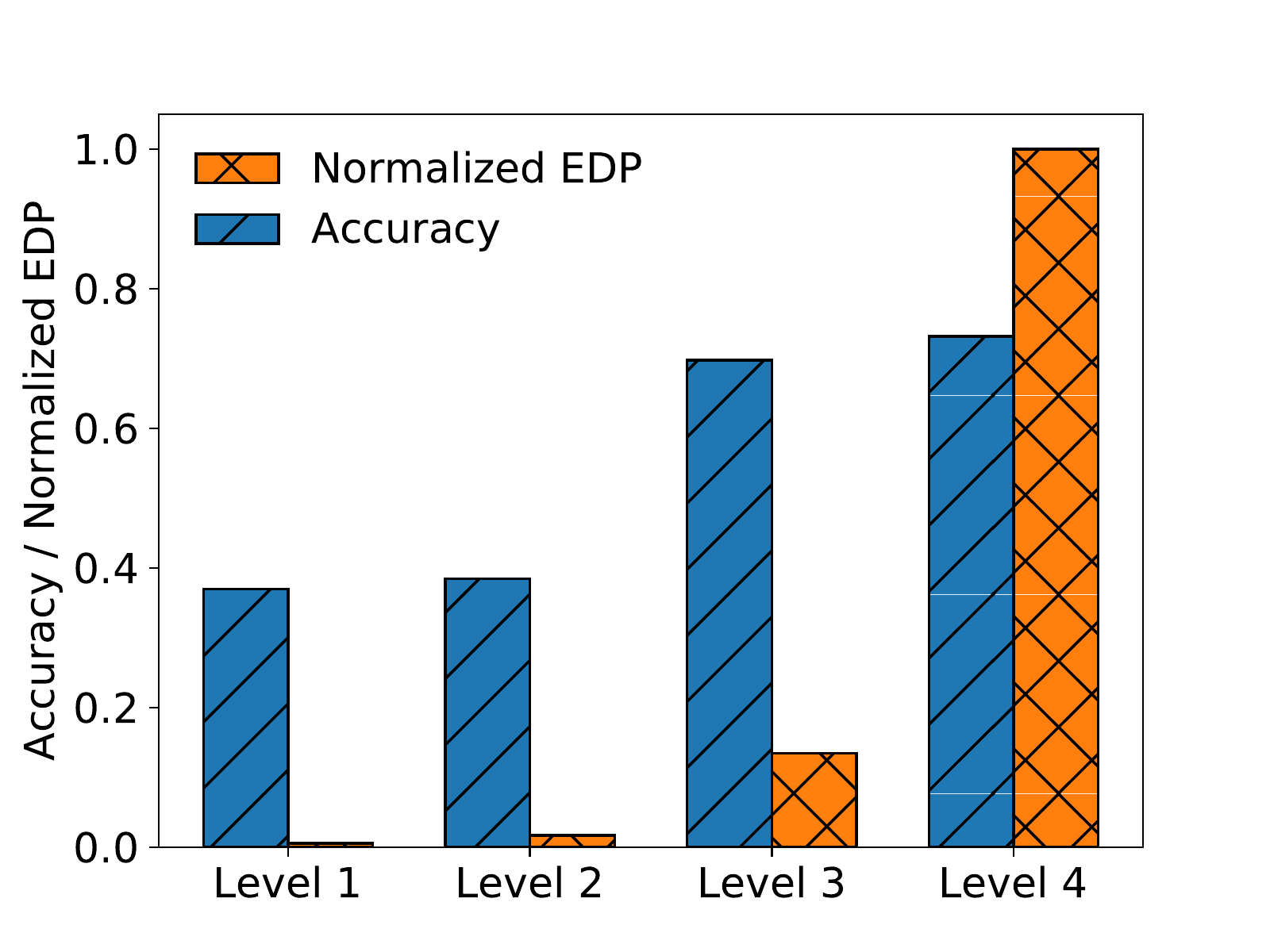}
%     \caption{c2f b}
	\centering
    \small (d) C2F Net D
    \end{minipage}
\caption{Accuracy and EDP when all predictions are made using an independently trained network at a particular level.} %(a) Results for three levels of C2F Net A. (b) Results for four levels of C2F Net B.}
\label{fig:base_perf}
% \vspace{-2.5ex}
\end{minipage}
\end{figure}

%\subsection{Base Network Performance} 

{\bf Accuracy and energy of networks at different levels.} We train and test the networks at different levels that are part of C2F Net. Specifically, we make all the predictions using a single network for each level of C2F Net. This experiment will characterize the accuracy and energy trade-off achieved by different networks (coarsest to finest) that are part of C2F Net. Figure~\ref{fig:base_perf} shows the accuracy and energy metrics for networks at different levels for all the C2F Nets A, B, C and D. {\bf C2F Net A:} From level 3 to level 2, we can see that accuracy drops by 5\% and EDP reduces by 50\%. Similarly, from level 2 to level 1, accuracy drops by 15\% and EDP reduces by 30\% w.r.t level 3. {\bf C2F Net B:} We achieve an accuracy of 92\% when we train and test the base network architecture B (i.e., level 3 network) on CIFAR10 data. We see a 7\% drop in accuracy and 85\% gain in EDP when we move from level 3 to level 2. {\bf C2F Net C and D:} We see the trend to be similar to Net A and B respectively. However, we observe higher accuracies with Net C because MNIST data set is simple and obtain lower accuracies in NET D because CIFAR100 data set is relatively complex.

In summary, we see a significant gain in energy with relatively small loss in accuracy when we move from finest to coarsest network in C2F Nets. This corroborates our hypothesis that we can save significant amount of energy if we are able to select coarser networks for large fraction of easy images and complex networks for hard images that are relatively small. 
% \vspace{-1ex}

\begin{figure}[h]
\begin{minipage}{\linewidth}
\begin{minipage}{0.495\linewidth}
    \includegraphics[trim={0 0 45 40},clip, width={\linewidth}]{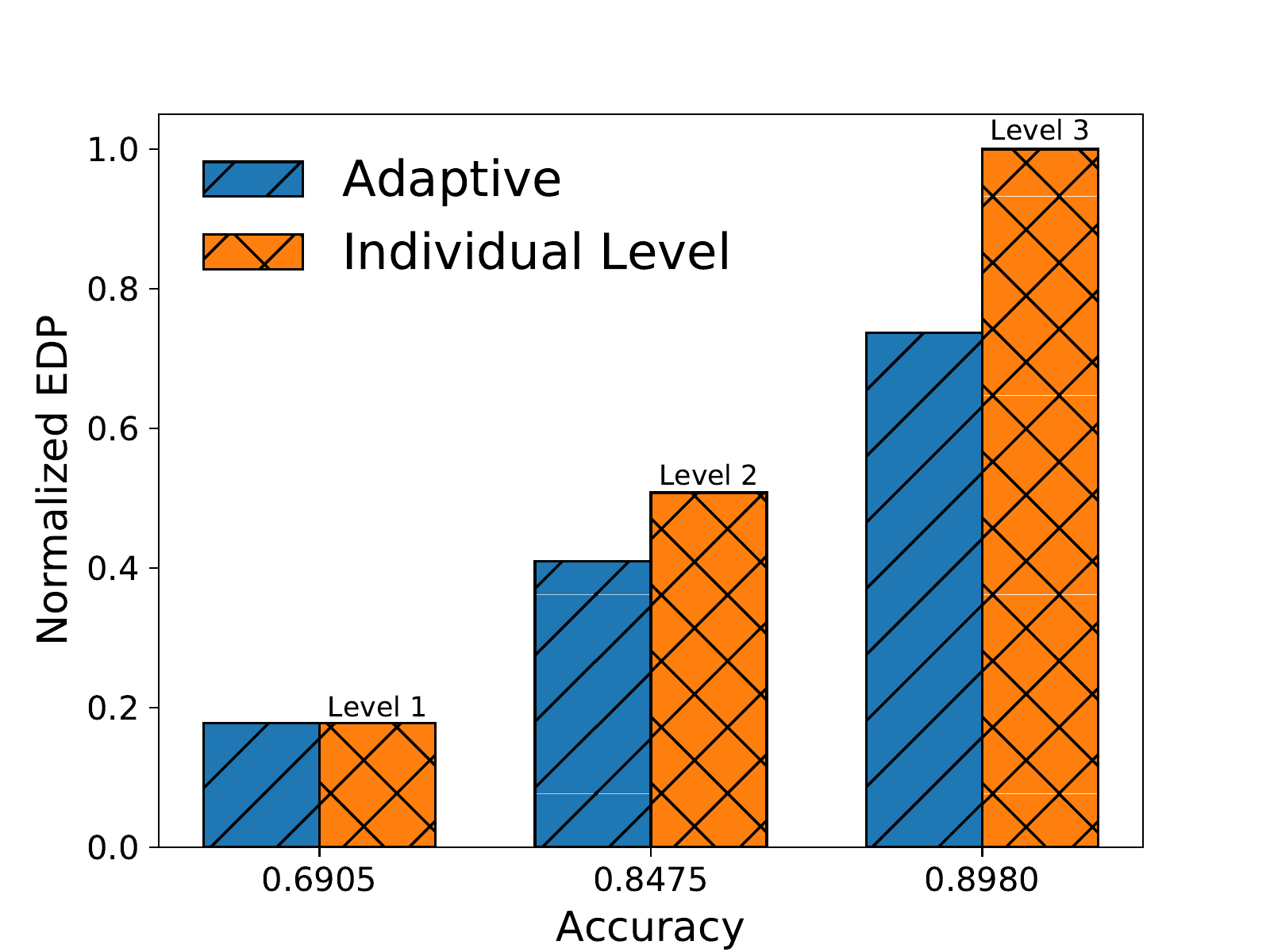}
    \centering
    \small (a) C2F Net A
\end{minipage}
%      \hspace{\fill} % note: no blank line here
    \begin{minipage}{0.495\linewidth}
    \includegraphics[trim={0 0 45 40},clip, width=\linewidth]{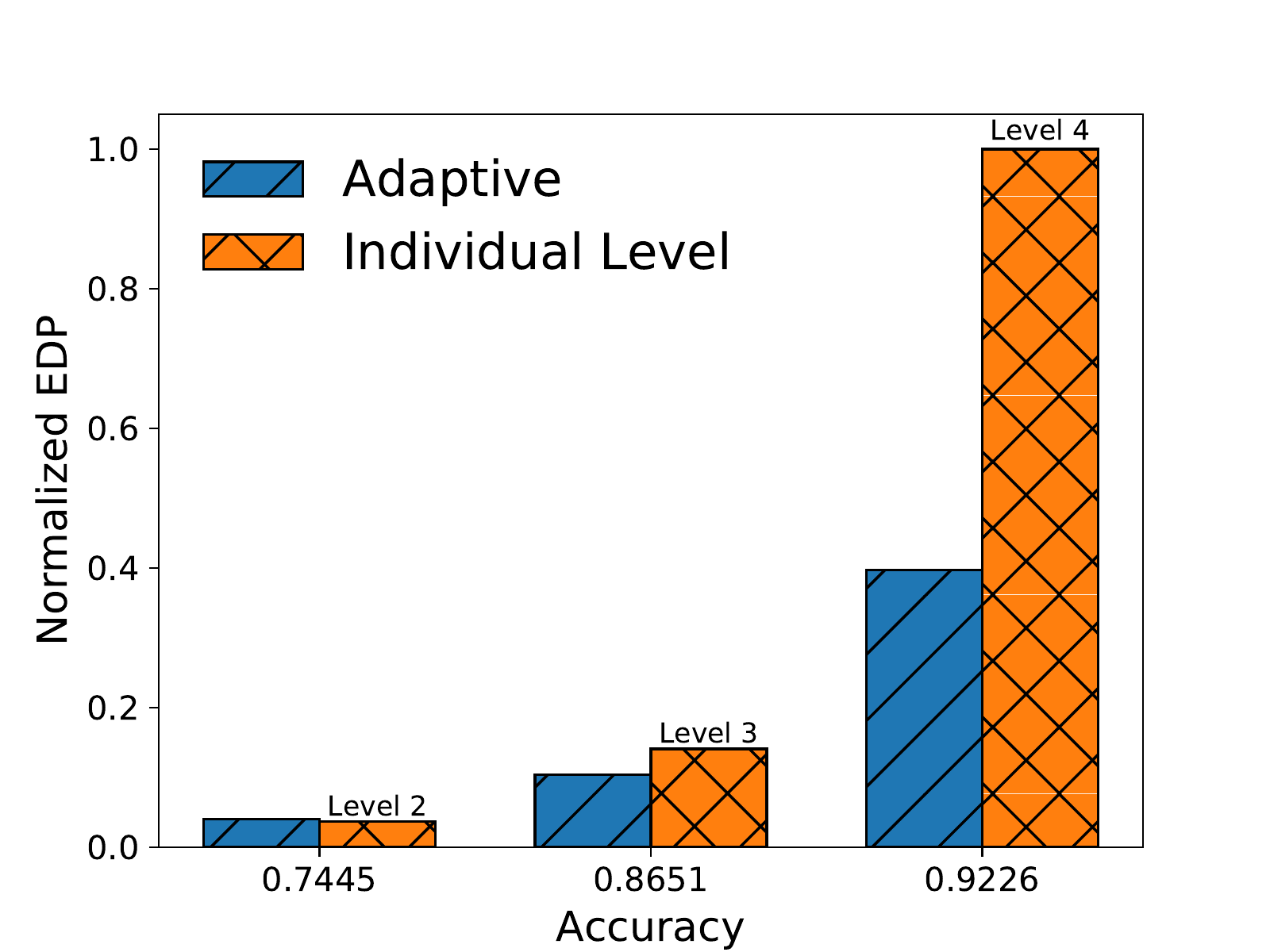}
	\centering
    \small (b) C2F Net B
    \end{minipage}
    \begin{minipage}{0.495\linewidth}
    \includegraphics[trim={0 0 45 40},clip, width=\linewidth]{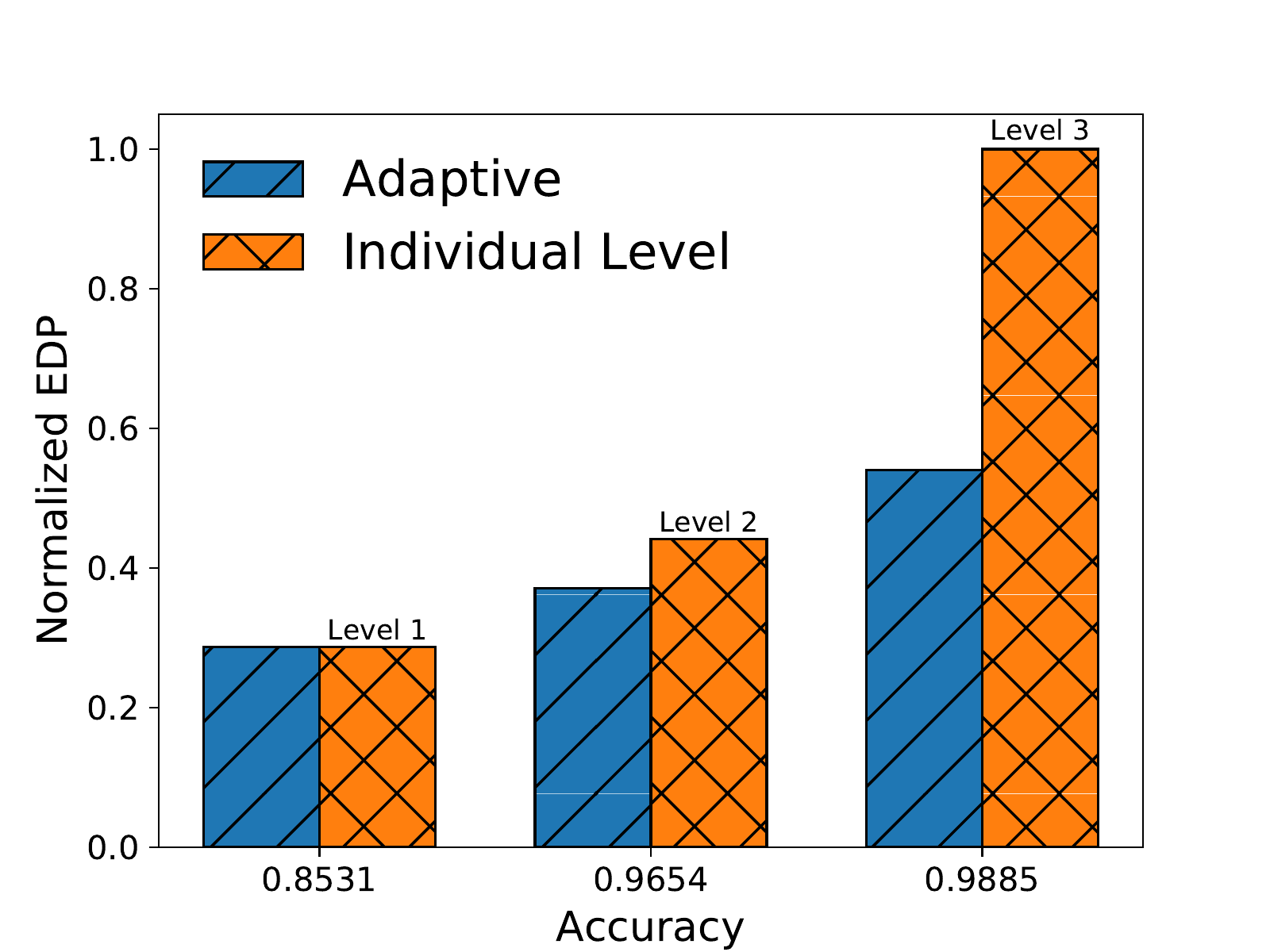}
%     \caption{c2f b}
	\centering
    \small (c) C2F Net C
    \end{minipage}
    \begin{minipage}{0.495\linewidth}
    \includegraphics[trim={0 0 45 40},clip, width=\linewidth]{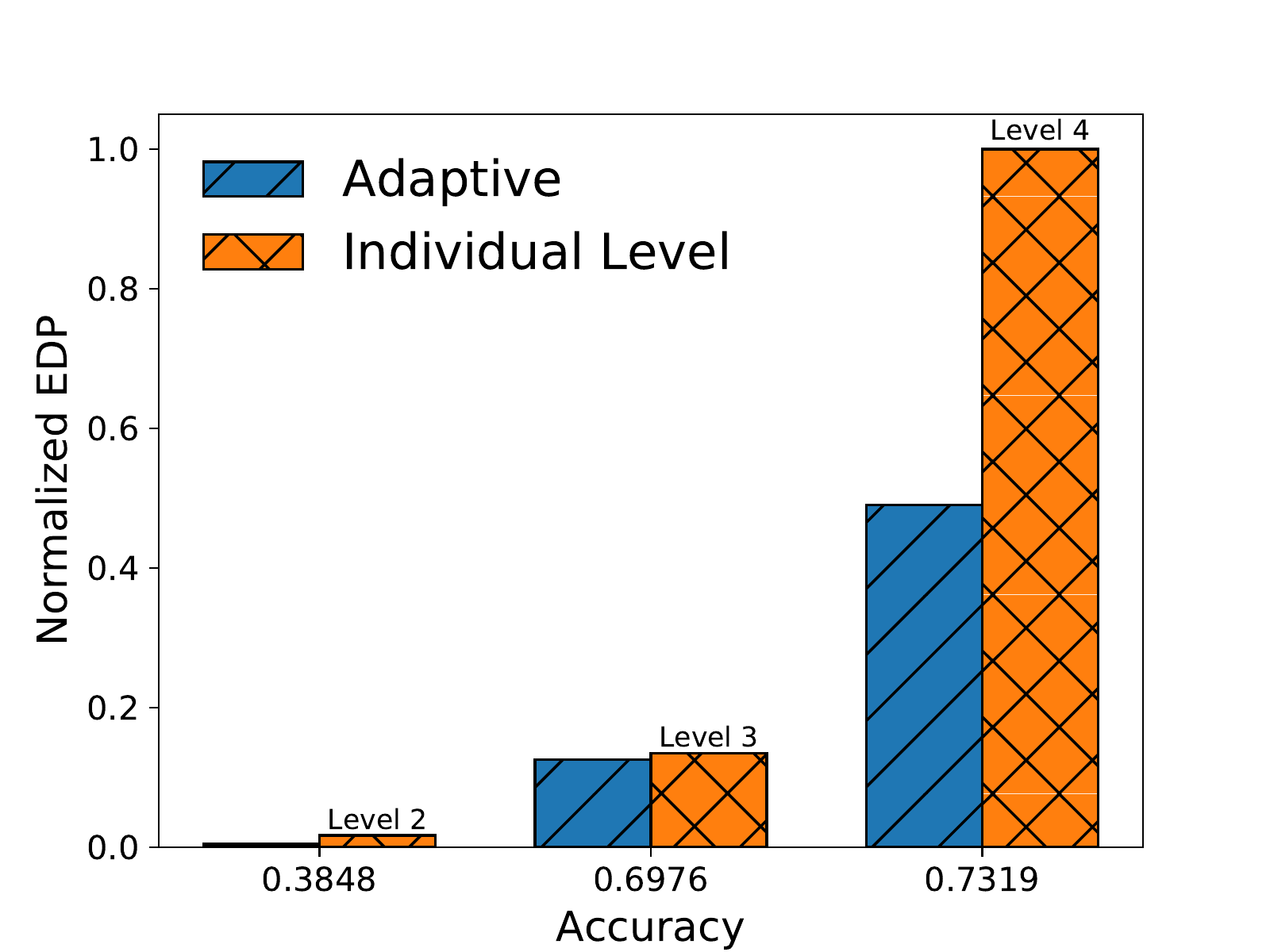}
%     \caption{c2f b}
	\centering
    \small (d) C2F Net D
    \end{minipage}
\caption{Energy consumption of adaptive C2F Nets optimized to achieve the accuracy of networks at different levels. Energy values are normalized w.r.t the most complex network.}

\label{fig:base_vs_pareto}
% \vspace{-2.5ex}
\end{minipage}
\end{figure}

{\bf Optimized C2F Net for accuracy of different networks.} We saw that fixed networks take significantly more energy for small improvement in accuracy. On the other hand, optimized C2F Nets can potentially reduce the energy consumption to achieve the same amount of accuracy by performing input-specific adaptive inference. To test the effectiveness of adaptive C2F Nets, we trained C2F nets with different $\lambda$ (trade-off) values to find the configurations that achieve the same accuracies as networks at different levels (Figure \ref{fig:base_vs_pareto}). 

{\bf C2F Net A:} Adaptive C2F Net improves the EDP by 26\% and 20\% to achieve the same accuracies when compared to the base networks at level 2 (89.8\% accuracy) and level 1 (85\% accuracy). 

{\bf C2F Net B:} We see significantly more improvement in energy gains when compared to C2F Net A. For example, EDP improves by 60\% to achieve 92\% accuracy of level 3 (most complex network) and improves by 26\% for achieving 85\% accuracy of level 2 network.

{\bf C2F Net C and D:} Similar to the results for Net A and Net B, we observe that EDP improves by 46\% and 51\% respectively with respect to the base network.

In summary, our approach using adaptive C2F Nets can significantly reduce the energy consumption with negligible loss in accuracy when compared to base networks of different complexity that are part of C2F Net. Additionally, the energy gains increase significantly for complicated base networks (e.g. Net B and D).

% \vspace{1.0ex}

\begin{figure}[h]
\centering
\begin{minipage}{0.48\linewidth}
\includegraphics[trim={0 0 45 40},clip, width=\linewidth]{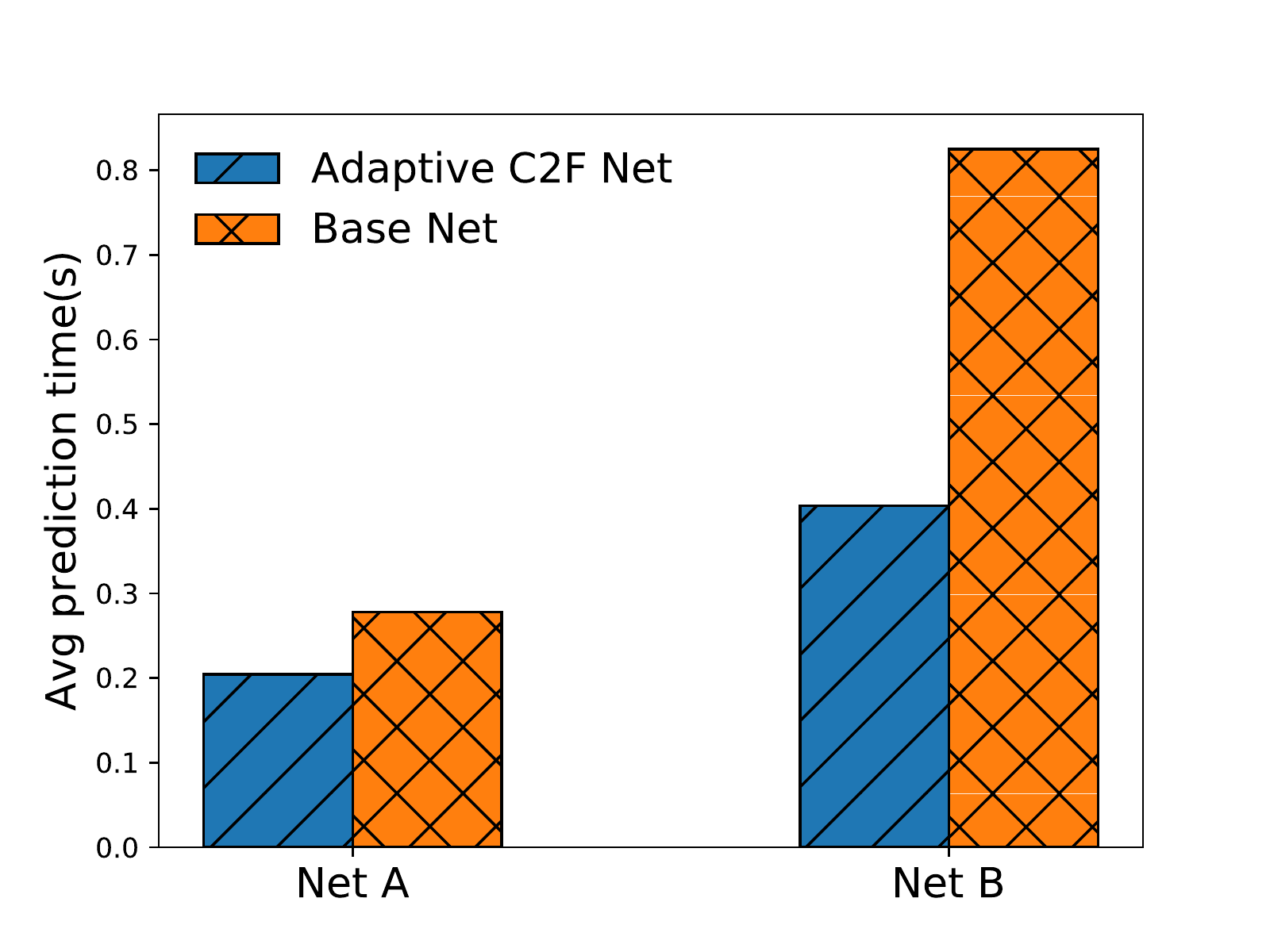}
\caption{Average prediction time in seconds for adaptive C2F Nets and finest (most complex) network. Adaptive C2F Nets are optimized to achieve the same accuracy as the finest network.}
\label{fig:c2f_avg_time_comp}
\end{minipage}
\begin{minipage}{0.010\linewidth}
\end{minipage}
% \vspace{3.0ex}
% \begin{figure}[h]
% \centering
\begin{minipage}{0.48\linewidth}
\includegraphics[trim={0 0 45 40},clip, width=\linewidth]{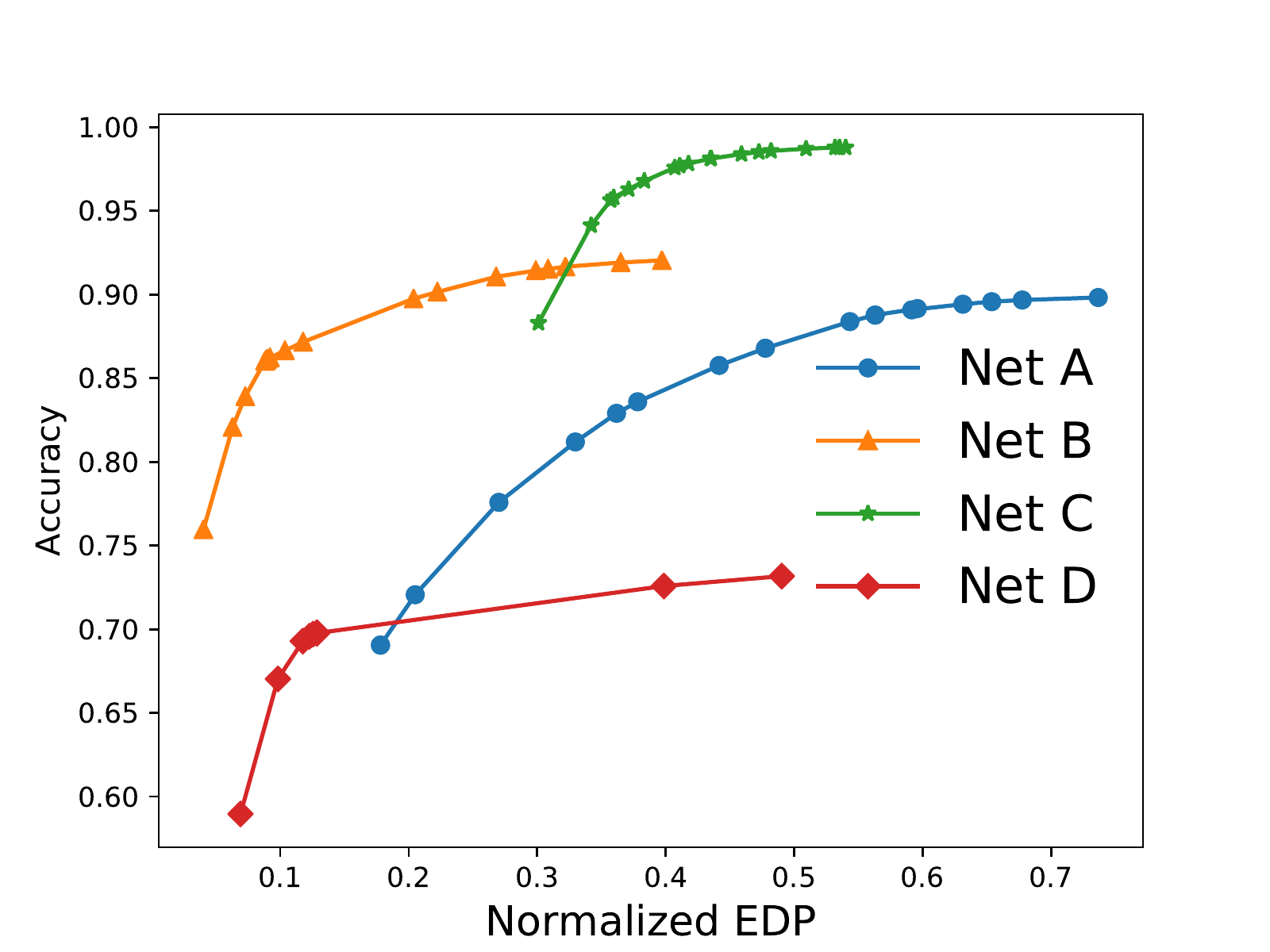}
\caption{Pareto-optimal curves obtained by varying $\lambda$ value (trade-off between energy and accuracy) for different C2F Nets: A (CIFAR10), B (CIFAR10), C (MNIST), D (CIFAR100).}
\label{fig:c2f_conf_max_prob_comp}
% \vspace{-2.5ex}
\end{minipage}
\end{figure}

{\bf Prediction time of adaptive C2F Net vs. finest network.} We compare the average execution time to make predictions using adaptive C2F Nets and the finest (most complex) network, where C2F Nets are optimized to achieve the same accuracy as the finest network. Figure~\ref{fig:c2f_avg_time_comp} shows these results. For network A, the average prediction time of base network and adaptive C2F net are $\sim$0.27 secs and $\sim$0.20 secs respectively, i.e., 26\% reduction in prediction time. Similarly, for network B, the prediction time reduces from  $\sim$0.82 secs to  $\sim$0.40 secs leading to 52\% reduction. In summary, our adaptive C2F networks perform better in terms of prediction time to achieve the same accuracy as the base network. Additionally, the improvement is much more for complex networks similar to energy gain results. 

% \vspace{1.0ex}

{\bf Fine-grained analysis of adaptive C2F Nets.} We demonstrated that adaptive C2F nets can improve both energy and prediction time when compared to the finest network through the above presented results. We perform fine-grained analysis to understand how adaptive C2F nets achieve these gains. We present this analysis for C2F Net A noting that analysis for other C2F nets show similar trends. 

C2F Net A was optimized for achieving the same accuracy (89.8\%) as the finest network. The energy gains can be explained by understanding how many images out of the 10000 image testing set are classified at different levels. At level 1, 263 images are classified with 100\% accuracy. At level 2, 4911 images are classified with with 98.7\% accuracy. At level 3, 4826 images are classified with 80.2\% accuracy. Therefore, 2\%, 48\%, and 49\% of the total testing images are classified at levels 1, 2, and 3 respectively. From previous results, we see that the accuracy of 89.8\% is obtained using 76\% of the energy consumed by the finest network. Levels 1, 2, and 3 contribute 0.46\%, 24.5\% and 48.26\% to this 76\%  respectively. Similarly, for C2F Net B with four levels, we see 0, 1339, 5517, and 3144 images classified with accuracies of 0, 99.7\%, 97.17\% and 79.73\% and energy consumption of 0.0, 0.49\%, 7.7\% and 31.4\% respectively. In both cases, more than 50\% of the images were predicted by a level other than the last level. Additionally, accuracy of the lower levels is closer to 100\%. Therefore, we see huge improvements in energy and prediction time for adaptive C2F Nets.

\begin{figure}[h]
\begin{minipage}{\linewidth}
\centering
\begin{minipage}{0.49\linewidth}
    \includegraphics[width=\linewidth]{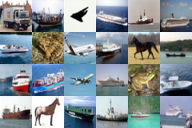}
    \centering
    \small (a) Images predicted at level 1 (coarsest network)
\end{minipage}
%     \hspace{\fill} % note: no blank line here
\begin{minipage}{0.49\linewidth}
    \includegraphics[width=\linewidth]{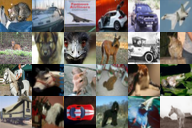}
    \centering
    \small (b) Images predicted at Level 3 (finest network)
\end{minipage}

\caption{Predicted images at different levels for C2F Net A} \label{fig:diff_lvl_img_pred}
% \vspace{-2.5ex}
\end{minipage}
\end{figure}

{\bf Qualitative results.} Figure~\ref{fig:diff_lvl_img_pred} shows some sample images that are predicted at level 1 (coarsest) and level 3 (finest) using adaptive C2F Net A. We make following observations: 1) Images classified using level 1 are simpler with a single object and clear background; 2) Images classified using level 2 have overlapping or hidden objects with confusing backgrounds. These results demonstrate how adaptive C2F nets use simpler classifiers for easy images and complex classifiers for hard images.

% \vspace{3.0ex}

\begin{figure}[h]
\begin{minipage}{\linewidth}
\centering
\begin{minipage}{0.48\linewidth}
    \includegraphics[trim={0 0 45 40},clip, width=\linewidth]{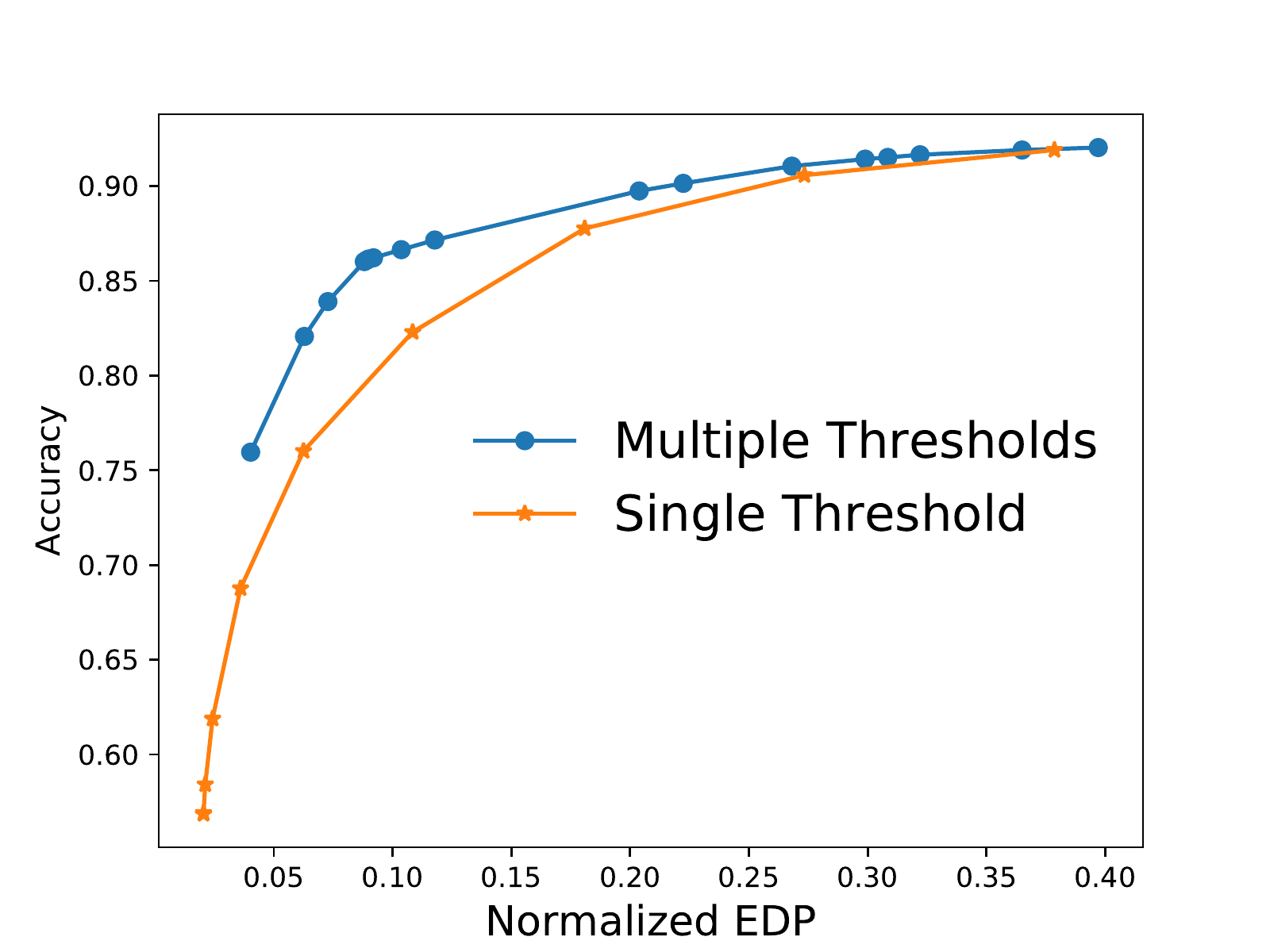}
    \centering
    \small (a) C2F Net B
    \end{minipage}
%     \hspace{\fill} % note: no blank line here
    \begin{minipage}{0.48\linewidth}
    \includegraphics[trim={0 0 45 40},clip, width=\linewidth]{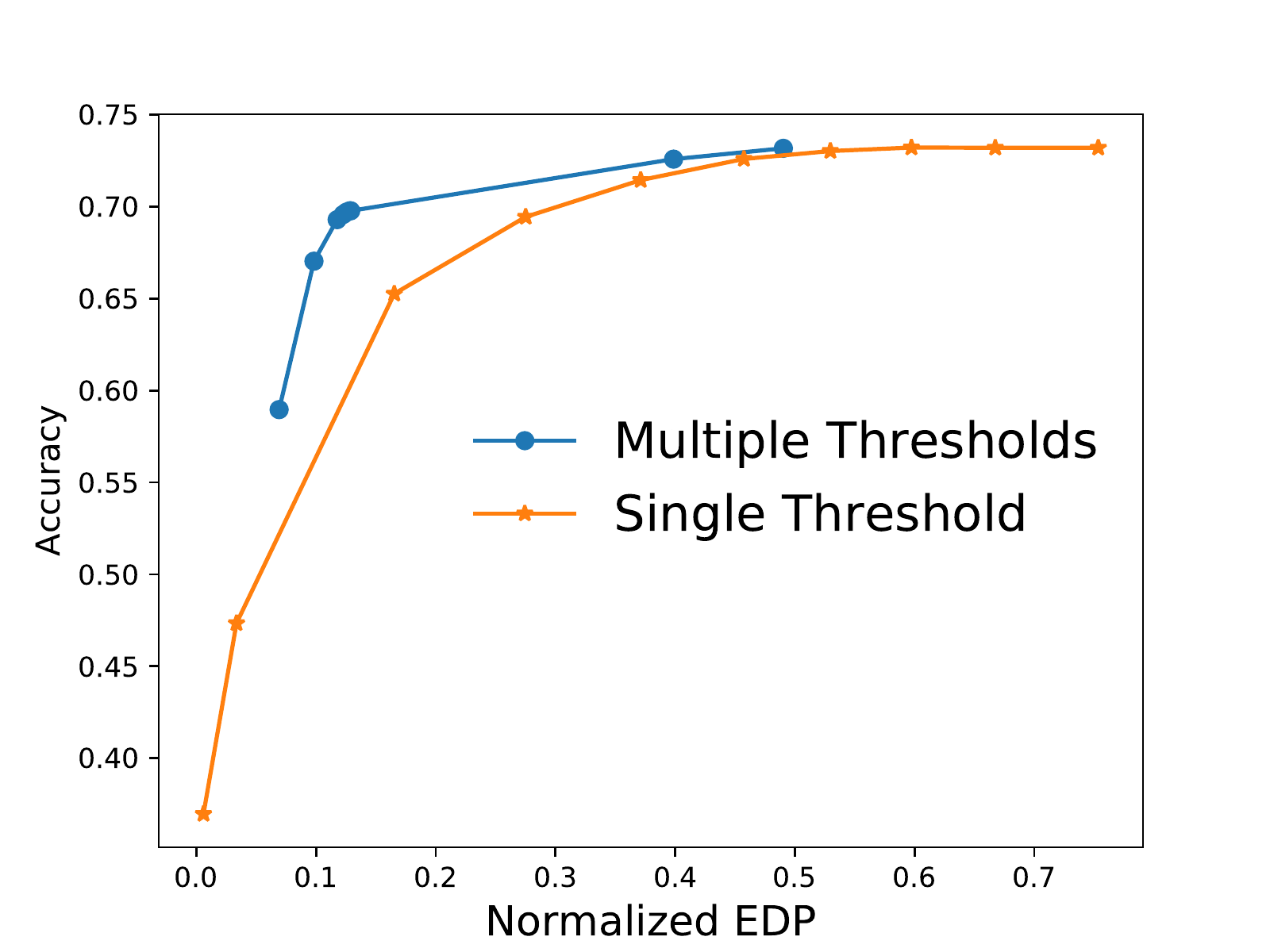}
%     \caption{c2f b}
	\centering
    \small (b) C2F Net D
    \end{minipage}

\caption{Comparison of pareto-optimal curves: single threshold for all levels vs. different thresholds for different levels.}
\label{fig:sgl_vs_mult_thresh}
% \vspace{-2.5ex}
\end{minipage}
\end{figure}

{\bf Single threshold vs. Multiple threshold.} To measure the quantitative difference between using single threshold for all levels of C2F net (as in CDL \cite{CondDeepLearn}) and different thresholds for all levels, we compare their pareto-optimal curves. Figure~\ref{fig:sgl_vs_mult_thresh} shows the comparative results for adaptive C2F networks B and D. We make the following observations. When the number of levels are more, different thresholds for different levels achieve better pareto-optimal solutions when compared to the setting with single threshold for all levels. Our experimental results on CIFAR10 (Net B) and CIFAR100 (Net D) clearly show this difference. 
We also present the pareto-optimal solutions we obtained for all adaptive C2F Nets A, B, C and D in Figure~\ref{fig:c2f_conf_max_prob_comp} for completeness. 
%We can see smooth variation in energy and accuracy trade-off, which can be very useful to deploy adaptive C2F nets in different embedded systems with varying energy/accuracy constraints.

%Figure~\ref{fig:c2f_avg_time_comp_conf_max_prob_comp} (b) shows the pareto-optimal curve for adaptive C2F networks by varying $\lambda$ value to get different energy-accuracy trade-offs. We repeat this experiment using all three confidence types, namely, maximum probability, margin, and entropy. We make the following observations. First, we get almost same pareto-optimal curves with different confidence types. This demonstrates the robustness of our Bayesian Optimization approach in finding the appropriate confidence threshold values irrespective of their type. Second, we can see smooth variation in energy and accuracy trade-off, which can be very useful to deploy C2F nets over different embedded devices with varying energy constraints.

{\bf Behavior of optimized C2F Nets with different $\lambda$.} We can vary the trade-off parameter to obtain optimized C2F nets for different accuracy and energy trade-offs. When $\lambda$ is close to zero, there is more emphasis on saving energy without paying attention to accuracy. Similarly, when $\lambda$ is close to one, adaptive C2F nets try to achieve best possible accuracy by minimizing the overall energy consumption. %In earlier results, we already discussed how accuracy and energy consumption varies with different values of $\lambda$. 

\begin{figure}[h]
\begin{minipage}{\linewidth}
\centering
\begin{minipage}{0.75\linewidth}
    \includegraphics[trim={0 0 45 0},clip, width=\columnwidth]{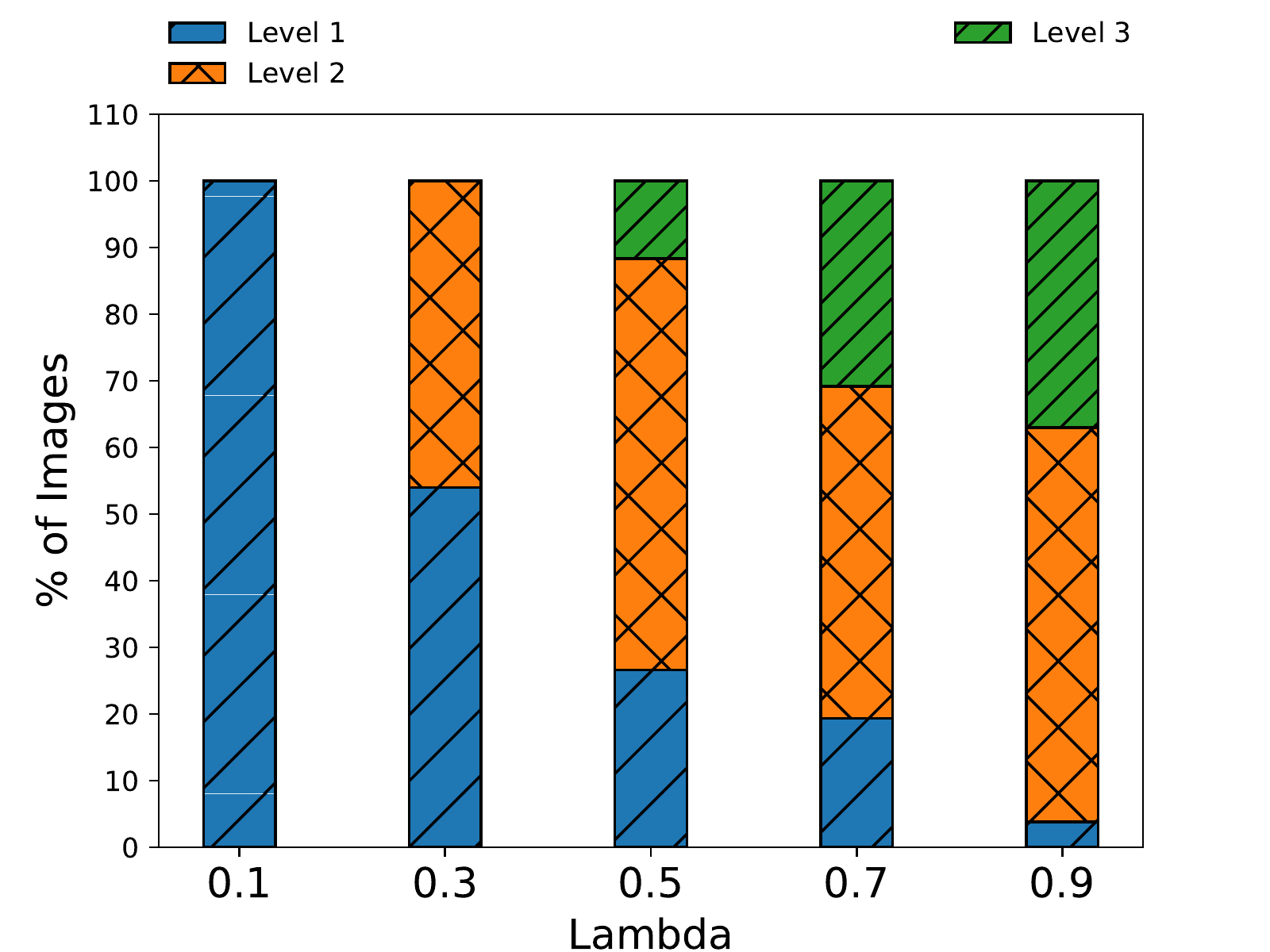}
    \centering
    \small (a) Num Images across Levels
\end{minipage}
%     \hspace{\fill} % note: no blank line here
    \begin{minipage}{0.75\linewidth}
    \includegraphics[trim={0 0 45 0},clip, width=\linewidth]{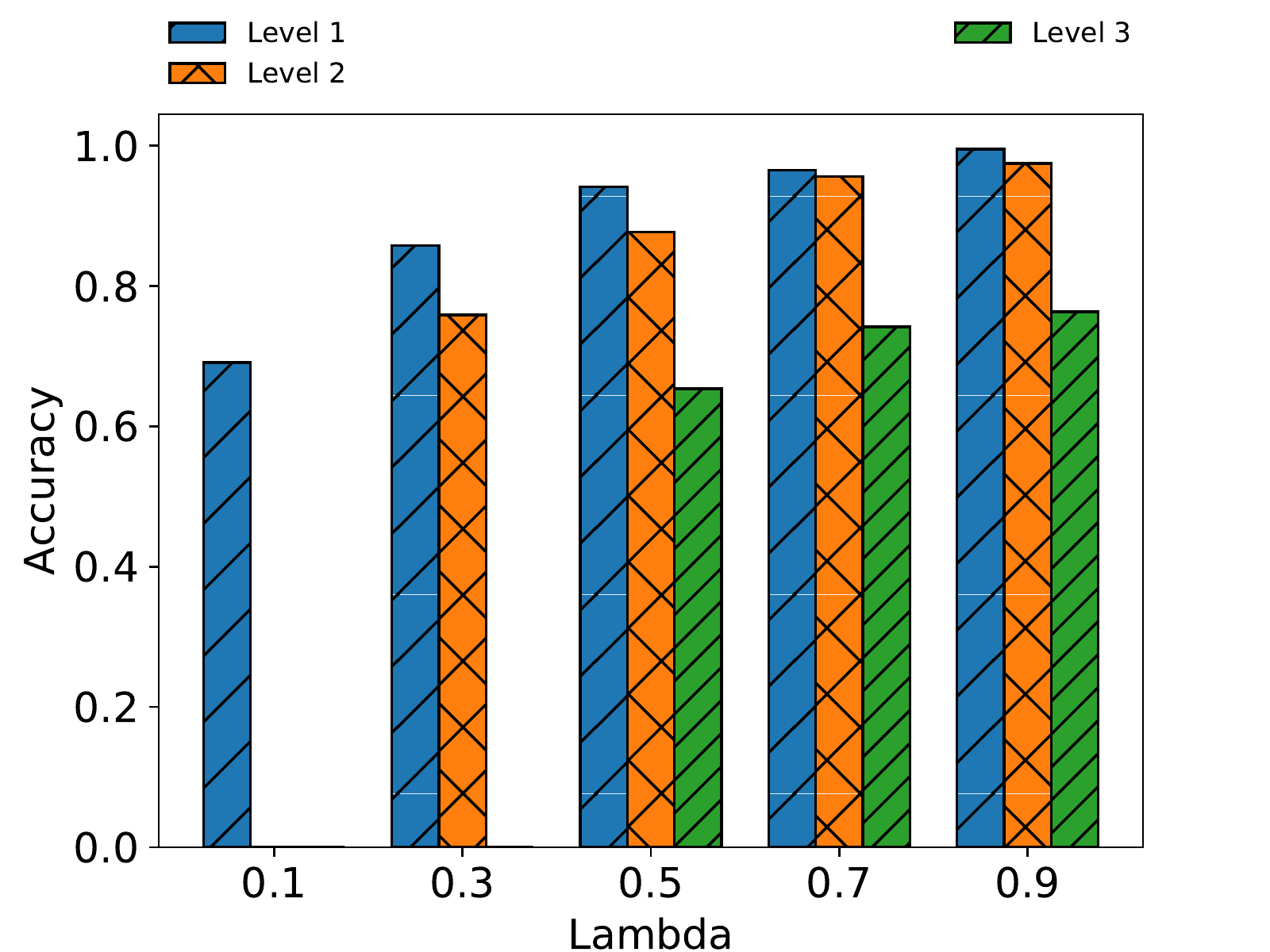}
%     \caption{c2f b}
	\centering
    \small (b) Accuracy across Levels
    \end{minipage}
    \begin{minipage}{0.75\linewidth}
    \includegraphics[trim={0 0 45 40},clip, width=\linewidth]{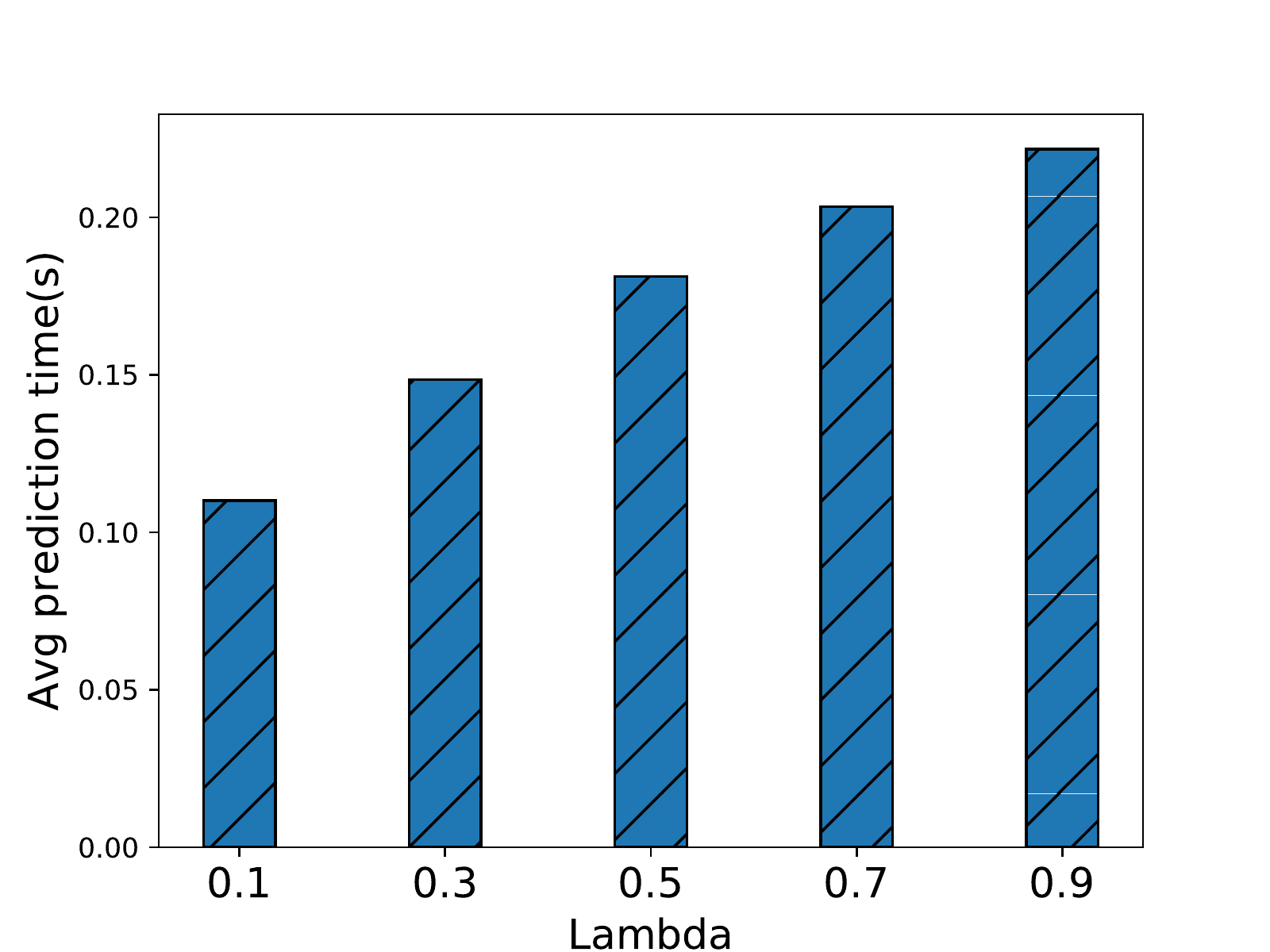}
%     \caption{c2f b}
	\centering
    \small (c) Prediction time
    \end{minipage}
\caption{Results for optimized C2F Net A by varying $\lambda$ values.} %(a) Distribution of images predicted at different levels, (b) Prediction accuracy at different levels, and (c) Average prediction time.}
%Effect of lambda on C2F Net A. As lambda varies we how the number of images predicted at each level [Top] varies, how the prediction accuracy at each level [Middle] varies and how the average prediction time of the entire C2F net [Bottom] varies.} 
\label{fig:variation_with_lambda}
% \vspace{-0.5ex}
\end{minipage}
\end{figure}

We study the behavior of optimized C2F nets with different values of $\lambda$ in terms of the number of images predicted at different levels, the prediction accuracy at different levels, and the average prediction time. Figure~\ref{fig:variation_with_lambda} shows these results for C2F Net A noting that results are similar for other C2F nets. 

%\vspace{-3.5ex}
We make following observations from Figure~\ref{fig:variation_with_lambda}(a). When $\lambda$=0, all the images are predicted at level 1 as coarsest network consumes the least energy. As the $\lambda$ value increases, the number of images predicted by level 2 and level 3 slowly increases to achieve higher accuracy. Note that, even with the highest $\lambda$ value, not all images are predicted by level 2 (finest network) to optimize the energy consumption to achieve high accuracy.

From Figure~\ref{fig:variation_with_lambda}(b), we can see that as $\lambda$ increases, the prediction accuracy of level 1 and level 2 increases proportionally and reach almost 100\% accuracy. The overall accuracy of C2F net is primarily determined by the accuracy of level 2 classifier.

From Figure~\ref{fig:variation_with_lambda}(c), we observe that the average prediction time gradually increases with $\lambda$, which is explained by more and more images getting predicted at higher levels. When compared to the base (finest) network, the prediction time at $\lambda$=0 is 63\% lesser and at highest $\lambda$ value, we get around 26\% gain. Therefore, in scenarios where we require real-time predictions, we can optimize the accuracy of C2F nets for the target time constraints.

% \vspace{-1.5ex}
\section{Conclusions and Future Work}

Motivated by the challenges associated with performing inference using deep neural networks on resource-constrained embedded systems, we studied a co-design approach based on the formalism of coarse-to-fine network (C2F Net) that allows us to employ classifiers of varying complexity depending on the hardness of input examples. Our proposed optimization algorithm to automatically configure a C2F Net for a specified trade-off between accuracy and energy consumption on a target hardware platform is very effective. Results on four different C2F Nets for image classification show that using optimized C2F Net we can significantly reduce the overall energy with no loss in accuracy when compared to a baseline solution, where predictions for all input examples are made using the most complex network. Though we have demonstrated the method for four nets this idea could be extended to other architectures like MobileNets \cite{MobileNets}. Future directions include evaluation of our framework on deep networks such as MobileNets \cite{MobileNets}, studying automated approaches for C2F nets design, heterogeneous architectures for embedded systems in the context of inference using deep networks, machine learning based approaches for software and hardware co-design using C2F Nets, and dynamic power management to further improve the overall energy-efficiency. 

%\noindent {\bf Funding Acknowledgements.} This work was supported in part by the Army Research Office grant W911NF-17-1-0485 from their computing sciences program.

% \vspace{-2.5ex}
% \input{references}
% \bibliographystyle{ACM-Reference-Format}
\bibliographystyle{IEEEtran}
\bibliography{references}
\vspace{-5.5ex}
\begin{IEEEbiography}[{\includegraphics[width=1in,height=1.5in,clip,keepaspectratio]{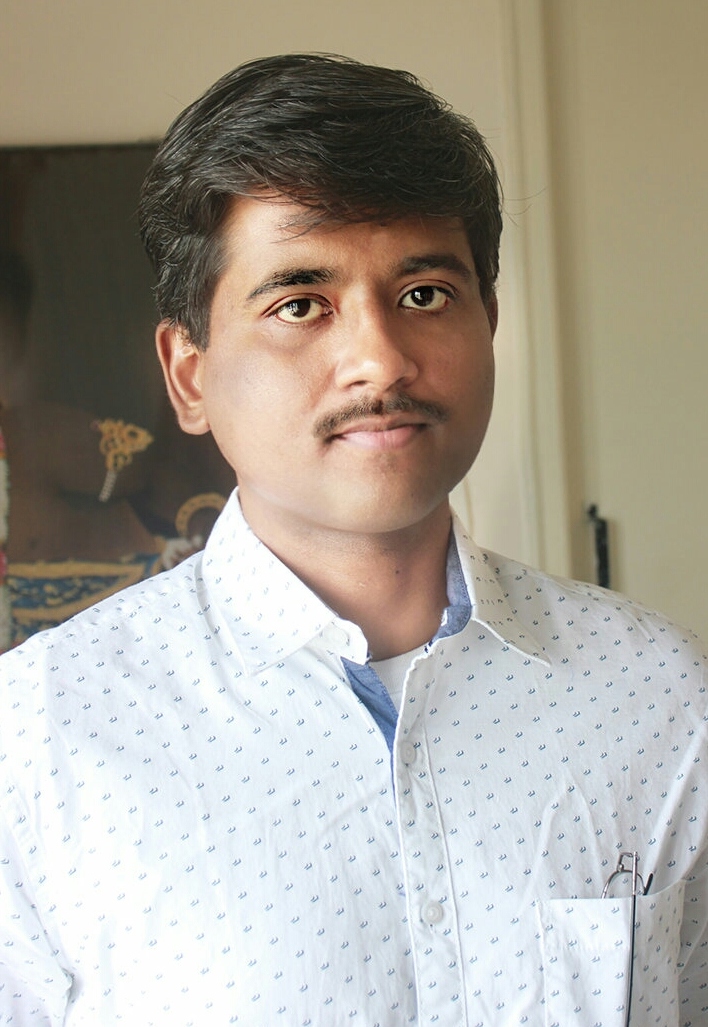}}]{Nitthilan Kannappan Jayakodi}(S’17) is currently pursuing his Ph.D. degree with the Electrical and Computer Engineering Department, Washington State University, Pullman, WA, USA. His current research interests are at the intersection of machine learning and electronic design automation.
%His current research interest includes low-power network-on-chip design.
\end{IEEEbiography}
\vspace{-5.5ex}

\begin{IEEEbiography}[{\includegraphics[width=1in,height=1.5in,clip,keepaspectratio]{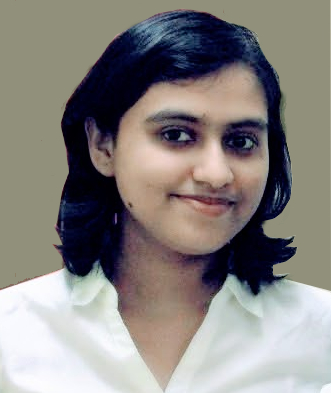}}]{Anwesha Chatterjee}
(S’17) is currently pursuing her Ph.D. degree with the Electrical and Computer Engineering Department, Washington State University, Pullman, WA, USA. Her current research interest includes low power VLSI solutions, many-core chip design and machine learning.
\end{IEEEbiography}
\vspace{-5.5ex}

\begin{IEEEbiography}[{\includegraphics[width=1in,height=1.5in,clip,keepaspectratio]{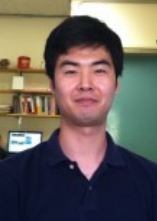}}]{Wonje Choi}
(S’17) got his Ph.D. degree (2018) from the Electrical and Computer Engineering Department, Washington State University, Pullman, WA, USA. He is currently working as a senior design engineer at AMD. His current research interest includes low-power network-on-chip design and dynamic power management for heterogeneous many-core systems.
\end{IEEEbiography}
\vspace{-2.5ex}

\begin{IEEEbiography}[{\includegraphics[width=1in,height=1.5in,clip,keepaspectratio]{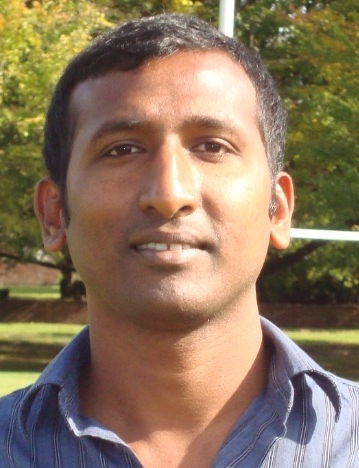}}]{Janardhan Rao Doppa}
(M’14) received the Ph.D. degree from Oregon State University, Corvallis, OR, USA. 
He is an Assistant Professor with the school of Electrical Engineering and Computer Science, Washington State University, Pullman, WA, USA. His current research is at the intersection of machine learning and electronic design automation. Dr. Doppa was a recipient of the Outstanding Paper Award for the research on structured prediction at the AAAI 2013 conference.\end{IEEEbiography}
\vspace{-2.5ex}

\begin{IEEEbiography}[{\includegraphics[width=1in,height=1.5in,clip,keepaspectratio]{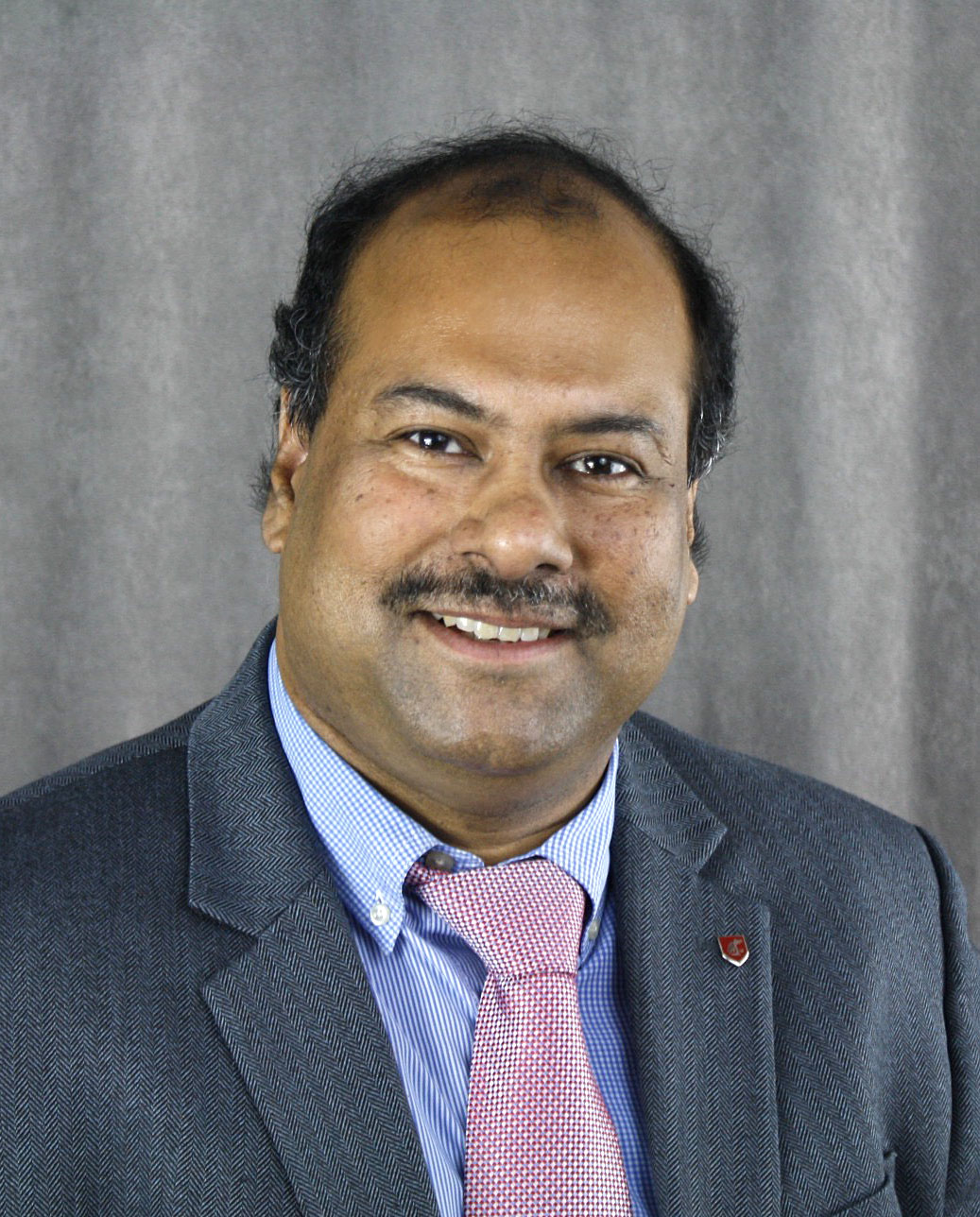}}]{Partha Pratim Pande}
(SM’11) received the M.S. degree in computer science from the National University of Singapore, Singapore, in 2002, and the Ph.D. degree in ECE from the University of British Columbia, Vancouver, BC, Canada, in 2005. He is a Professor and holds the Boeing Centennial Chair in computer engineering with the School of Electrical Engineering and Computer Science, Washington State University, Pullman, WA, USA.\end{IEEEbiography}

\end{document}